\documentclass[]{template}
\newcommand{\MethodName}{SpatialASM}

\usepackage[utf8]{inputenc}             
\usepackage[T1]{fontenc}                
\usepackage{url}                        
\usepackage{booktabs}                   
\usepackage{multirow}
\usepackage{colortbl}
\usepackage{multicol}
\usepackage{amsfonts}                   
\usepackage{nicefrac}                   
\usepackage{microtype}                  
\usepackage[dvipsnames]{xcolor}         

\usepackage{latexsym}

\usepackage{graphicx}
\usepackage{float}
\usepackage{subcaption}
\usepackage{wrapfig}
\usepackage{lipsum}

\usepackage{bm}

\usepackage{tabularx} 
\usepackage{ragged2e} 
\newcolumntype{L}{>{\RaggedRight\hangafter=1\hangindent=0em}X}

\usepackage{enumitem}

\usepackage{amsmath}
\usepackage{amssymb}
\usepackage{mathtools}
\usepackage{amsthm}

\setboolean{logo}{true}    

\usepackage[linesnumbered,ruled,vlined]{algorithm2e}

\hypersetup{
    colorlinks=true,
    linkcolor=red,
    citecolor=Cerulean,
    filecolor=magenta,      
    urlcolor=magenta,
}

\usepackage[capitalize,noabbrev]{cleveref}
\crefname{section}{§}{§§}
\Crefname{section}{§}{§§}

\usepackage{calligra}
\DeclareMathAlphabet{\mathcalligra}{T1}{calligra}{m}{n}

\usepackage{pifont}

\theoremstyle{plain}

\theoremstyle{definition}

\theoremstyle{remark}

\renewcommand{\paragraph}[1]{\vspace{1mm}\noindent\textbf{#1}}

\DeclareCaptionLabelFormat{cont}{#1~#2\alph{ContinuedFloat}}
\usepackage[most]{tcolorbox}
\tcbset{
  promptbox/.style={
    top=10pt,
    colback=lightgray!20,
    colframe=Black,
    colbacktitle=NavyBlue,
    enhanced,
    center,
    attach boxed title to top center={yshift=-0.1in,xshift=0.0in},
    boxed title style={boxrule=0pt,colframe=white,},
  }
}
\newtcolorbox{promptbox}[2][]{promptbox, title=#2,#1}
\tcbset{
  takeawaybox/.style={
    top=10pt,
    colback=lightgray!20,
    colframe=Black,
    colbacktitle=BurntOrange,
    enhanced,
    center,
    attach boxed title to top center={yshift=-0.1in,xshift=0.0in},
    boxed title style={boxrule=0pt,colframe=white,},
  }
}
\newtcolorbox{takeawaybox}[2][]{takeawaybox, title=#2,#1}
\tcbset{
  observationbox/.style={
    top=10pt,
    colback=lightgray!20,
    colframe=Black,
    colbacktitle=YellowGreen,
    enhanced,
    center,
    attach boxed title to top center={yshift=-0.1in,xshift=0.0in},
    boxed title style={boxrule=0pt,colframe=white,},
  }
}
\newtcolorbox{observationbox}[2][]{observationbox, title=#2,#1}

\usepackage{xspace}

\newcommand\blfootnote[1]{%
  \begingroup
  \renewcommand\thefootnote{}\footnote{#1}%
  \addtocounter{footnote}{-1}%
  \endgroup
}

\usepackage{CJK}

\title{SpatialASM: Internalizing Active Mentalization for Spatial Reasoning}

\author[$*$1,2]{Yuhong Liu}
\author[$*$1,2]{Shenxi Wu}
\author[2,3]{Junyi Chi}
\author[1,2]{Beichen Zhang}
\author[2,3]{Xuanlang Dai}
\author[$\dagger$4]{Jiaqi Wang}
\author[$\dagger$2]{Yuhang Zang}
\author[$\dagger$1,2,5]{Dahua Lin}

\affil[1]{The Chinese University of Hong Kong}
\affil[2]{Shanghai AI Laboratory}
\affil[3]{Fudan University}
\affil[4]{Shanghai Innovation Institute}
\affil[5]{CPII under InnoHK}

\begin{abstract}

When solving complex spatial reasoning tasks, ordinary humans naturally revisit the scenes from task-relevant viewpoints and mentally reconstruct the spatial layouts from those viewpoints. But can Large Vision-Language Models (LVLMs) proactively think in the same way? We term this ability \textbf{Active Spatial Mentalization (ASM)}, an intrinsic component of visual intelligence that remains substantially underdeveloped in current LVLMs. To bridge this gap, we introduce \textbf{SpatialASM}, a scalable framework to internalize ASM into LVLMs through training-time spatial simulation. SpatialASM employs an allocentric novel-view generator to ground hypothetical viewpoints in visual evidence and produces two complementary forms of supervision: \textbf{i)} explicit ASM trajectories that teach LVLMs when and how to adaptively invoke view planning and imagination during spatial reasoning, and \textbf{ii)} automatically-constructed perspective-taking tasks that further strengthen viewpoint-conditioned scene imagination. 
On seven benchmarks spanning relatively perception-oriented spatial understanding to more challenging tasks requiring multi-image, cross-view and viewpoint-transformation reasoning, SpatialASM yields performance gain on LVLMs of four LVLM backbones in spatial reasoning while retaining their original general vision capabilities. The experimental results demonstrate that our approach improves average accuracy across the seven spatial benchmarks. And the synthesis of ASM trajectories provides a scalable pathway to alleviate the scarcity of spatial Chain-of-Thought (CoT) supervision for eliciting active viewpoint exploration and mental scene simulation.

\end{abstract}

\begin{document}

\blfootnote{$\dagger$ Corresponding authors: zangyuhang@pjlab.org.cn}
\blfootnote{$*$ Equal contribution. Code is at \url{https://github.com/InternLM/SpatialASM}}

\maketitle

\section{Introduction}

\begin{figure}[h]
\centering
\includegraphics[width=1.0\textwidth]{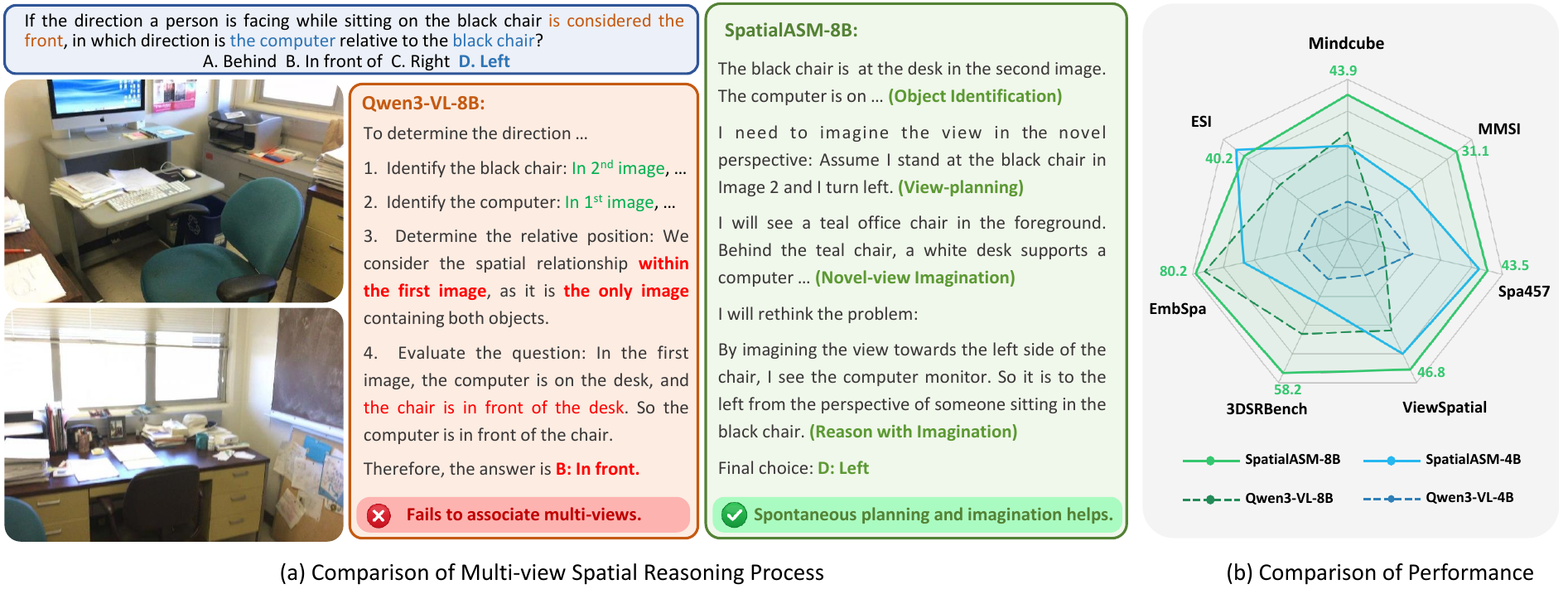}
\caption{\MethodName{} improves LVLM spatial reasoning by eliciting active exploration and imagination. (a) The qualitative example shows that the baseline LVLM (\textcolor[RGB]{255, 0, 0}{red})  fails to associate objects across views, whereas our model (\textcolor[RGB]{0, 185, 0}{green}) uses the internal viewpoint-planning skill and reasons over the imagined observation to reach the correct answer. (b) Quantitative experiments demonstrate that \MethodName{} delivers consistent performance gains on seven benchmarks.
}
\vspace{-5pt}
\label{fig:teaser}
\end{figure}

Spatial reasoning is a foundational capability for Large Vision-Language Models (LVLMs), underpinning applications from autonomous driving and embodied navigation to robotic manipulation~\citep{chen2024spatialvlm, cheng2024spatialrgpt, li2026spatialladder}.
While LVLMs achieve strong performance on many visual tasks~\citep{bai2026intern, liu2024mmbench, yang2026caprl++}, they still fail on complex spatial problems that involve viewpoint transformation, cross-view correspondence, and reasoning about unobserved regions~\citep{yang2026mmsi, yang2025thinking, fu2024blink, xu2026citycube}.
These tasks go beyond passively extracting relations from the given observations, as they require the model to reason about the three-dimensional scene from perspectives that the input images do not directly provide.

\begin{figure}[t]
\centering
\includegraphics[width=1.0\textwidth]{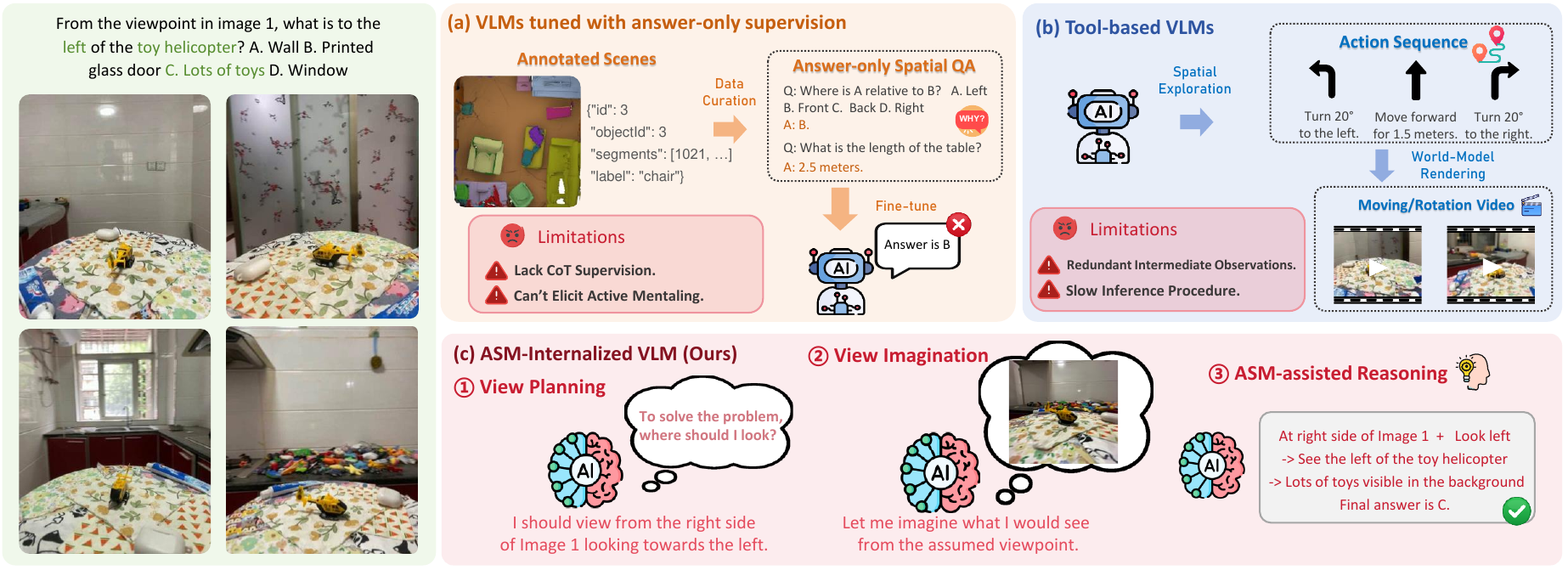}
\caption{(a) Prior data-centric approaches mostly provide answer-only supervision. (b) Reasoning with external calls of world models adds unnecessary intermediate observations and extra inference time. (c) \MethodName{} enables LVLMs to intrinsically produce plan-imagine-reason CoT.}
\label{fig:comparison}
\end{figure}

We argue that a core capability missing from current LVLMs is the ability to \textit{proactively} determine which alternative viewpoints would reveal useful evidence and to construct what the scene looks like from those viewpoints mentally~\citep{shepard1971mental}.
We term this joint capability \textbf{Active Spatial Mentalization (ASM)}, comprising two complementary skills: \textbf{view planning}, which selects task-relevant viewpoints given the question and current observations, and \textbf{view imagination}, which mentally constructs the spatial layout expected from the selected viewpoints (Fig.~\ref{fig:comparison}(c)).
Teaching ASM to LVLMs is difficult.
Recent data-centric efforts have expanded spatial supervision through large-scale datasets built from 3D annotations and videos~\citep{gao2026holi, liu2026openspatial, yang2026visual}, but these datasets provide answer-only QA pairs rather than Chain-of-Thought (CoT) trajectories that show \textit{how} to select informative viewpoints and imagine the corresponding observations (Fig.~\ref{fig:comparison}(a)).
Current LVLMs are unreliable at bootstrapping this supervision themselves, as they often produce incorrect viewpoint plans and hallucinated scenes when asked to generate ASM trajectories from scratch.

An alternative is to attach an external world model or spatial tool at inference time to generate supplementary observations~\citep{yang2026mindjourney, cao2026spatialdreamer, zhang2026think3d} (Fig.~\ref{fig:comparison}(b)).
While this provides useful visual evidence, it ties spatial reasoning to an external simulator at inference time, even when the LVLM learns when to invoke it~\citep{yu2026and, zhu2026thinking}. Tool distillation~\citep{zhou2026spatialcli, li2026omniview} removes this dependence, but targets perceptual tools such as depth and cognitive maps rather than viewpoint planning and view imagination.
Most such methods also adopt an egocentric interface that reaches a target viewpoint through a sequence of camera-centric actions, so each query requires a multi-step rollout~\citep{yang2026mindjourney, cao2026spatialdreamer}.
Closer to our setting, World2VLM~\citep{zhang2026world2vlm} uses a world model at training time, but it supervises egocentric view association without producing any reasoning trajectory.

We introduce \textbf{\MethodName{}}, a framework that equips LVLMs with ASM capabilities by using an allocentric novel-view generator (NVG) as a \textbf{training-time spatial simulation engine} rather than an inference-time tool (Fig.~\ref{fig:comparison}(c)).
Given a spatial QA sample, the NVG renders views from question-conditioned viewpoints planned by the LVLM, grounding hypothetical perspectives in concrete visual evidence.
\MethodName{} transforms these into two complementary forms of supervision: explicit ASM reasoning trajectories for supervised fine-tuning (SFT) and automatically verifiable perspective-taking tasks for reinforcement learning with GRPO~\citep{shao2024deepseekmath}.
The rendered views serve only as temporary scaffolding: in the final SFT data, they are replaced by text captions that describe the imagined observations, so the LVLM learns to construct spatial layouts internally from the original input images alone (Fig.~\ref{fig:teaser}(a)).

The design of \MethodName{} is guided by two key mechanisms (Fig.~\ref{fig:method}).
The \textbf{allocentric} viewpoint formulation specifies a target view by its location and direction in a single step, avoiding the multi-step rollouts of egocentric interfaces.
A \textbf{Geometry Perturbation (GeoP)} metric provides a model-based proxy for the geometric consistency of rendered views; together with an answer-consistency check, it admits only samples that pass pre-defined thresholds into the training data.

Across four backbones from the Qwen3-VL and Qwen3.5 series~\citep{bai2025qwen3, qwen3.5}, \MethodName{} improves average accuracy over seven spatial benchmarks (Tab.~\ref{tab:main}), with average gains ranging from $+1.2$ to $+3.5$ points (Fig.~\ref{fig:teaser}(b)).
The largest per-benchmark gains appear on tasks that demand spatial mental modeling and viewpoint transformation, reaching $+8.7$ on Spatial457-6d and $+5.2$ on MindCube.
General-VQA performance remains within $0.6$ points of the baselines across all four backbones.

Our contributions are threefold.
\textbf{1)} We formulate Active Spatial Mentalization as the joint capability of proactively planning task-relevant viewpoints and imagining corresponding observations, and argue that the absence of ASM is a key bottleneck of spatial reasoning in current LVLMs.
\textbf{2)} We introduce \MethodName{}, a scalable framework built on an allocentric novel-view generator and GeoP-based filtering that converts existing spatial QA data into explicit ASM trajectories for SFT and perspective-taking tasks for RL.
\textbf{3)} \MethodName{} improves average spatial accuracy on four LVLM backbones over seven benchmarks while keeping general-VQA performance within $0.6$ points.
\section{Related Work}

We thoroughly discuss more related work here in Appendix~\ref{Appendix:related}.

\textbf{Proactive spatial intelligence.} Current LVLMs achieve reasonable performance on static spatial perception~\citep{kamath2023s, wang2025site} but still struggle with tasks requiring cross-view association, spatial mental modeling, and active evidence acquisition~\citep{ray2024sat, zhang2026flatland, yang2026mmsi}.
ESI-Bench~\citep{hong2026esi} recasts spatial intelligence as a perception-action loop and shows that active exploration improves accuracy over passive observation, though its exploration relies on physical interaction with an embodied environment rather than internalized reasoning.
The two components of ASM have also been studied separately. For view imagination, APC~\citep{lee2025perspective} is a training-free pipeline that builds a 3D scene abstraction with vision foundation models and re-projects it to a perspective specified in the question. For view planning, recent work trains LVLMs to choose camera viewpoints, either to localize a given target view~\citep{wang2026planning} or to answer a question~\citep{koo2025toward}, but the chosen views are then rendered from a 3D environment rather than imagined.
GeoThinker~\citep{li2026thinking} internalizes a form of active reasoning at the feature-retrieval level by selectively querying geometric features, but such geometry-enhanced approaches~\citep{zheng2026learning, li2026thinking} require architecture-specific fusion modules that limit portability across LVLM architectures.
In contrast, ASM couples view planning and view imagination in a single question-conditioned reasoning trajectory that operates at the viewpoint and observation level and can be internalized through standard post-training without architectural modification.

\textbf{Spatial data synthesis.}
Many studies enhance spatial reasoning by synthesizing large-scale datasets (Fig.~\ref{fig:comparison}(a)). Early approaches build on public scene datasets (\textit{e.g.}, ScanNet~\citep{dai2017scannet}, ARKitScenes~\citep{baruch2021arkitscenes}) with 3D annotations~\citep{li2026spatialladder, ouyang2025spacer}, or employ expert models (\textit{e.g.}, segmentation, depth estimation) for data curation~\citep{chen2024spatialvlm, cheng2024spatialrgpt, deng2026internspatial}.
Recent data engines scale further: HoliSpatial~\citep{gao2026holi} renders scenes through 3D Gaussian Splatting, and VST~\citep{yang2026visual} uses simulators to generate novel observations.
Despite their effectiveness, the dominant supervision remains answer-only, providing the final answer but little guidance on \textit{how} to actively acquire or mentally construct missing evidence.
VST-R~\citep{yang2026visual} takes a step towards reasoning supervision with explicit step-by-step traces, but its trajectories teach the model to derive spatial conclusions from provided evidence rather than to plan informative viewpoints and imagine unobserved perspectives.
\MethodName{} addresses this gap by augmenting existing spatial QA pairs with ASM reasoning trajectories that couple question-conditioned view planning, viewpoint-conditioned imagination, and evidence-based reasoning.

\textbf{World-model-assisted spatial reasoning.} Drawing on generative world models~\citep{zhou2025stable, bruce2024genie, team2025hunyuanworld, chen2026definition, agarwal2026cosmos}, several approaches equip LVLMs with external spatial simulators during test-time reasoning (Fig.~\ref{fig:comparison}(b)).
MindJourney~\citep{yang2026mindjourney} and SpatialDreamer~\citep{cao2026spatialdreamer} use video-generation models to synthesize observations under egocentric motions in a multi-round think-imagine-reason loop, while Think3D~\citep{zhang2026think3d} allows the LVLM to manipulate a virtual camera inside a reconstructed 3D scene.
Astra~\citep{zhu2026thinking} and AVIC~\citep{yu2026and} further train the LVLM to decide when and how much to imagine, but the imagined views are still produced by an external simulator at inference time, and all of these methods reach a target viewpoint through egocentric action sequences.
A second line uses world models or renderers only at training time. World2VLM~\citep{zhang2026world2vlm} distills motion-conditioned view association from a world model, with trajectories sampled independently of any question and without reasoning traces; GeoWorld-VLM~\citep{gu2026geoworld} aligns the vision encoder with world-model features under a frozen LLM; and Imaginative Perception Tokens~\citep{bigverdi2026imaginative} train a unified model to generate latent views of a viewpoint specified in the question, supervised by simulator renders.
\MethodName{} belongs to this second line. Its difference lies not in using a world model at training time but in what is distilled: question-conditioned view planning and viewpoint-conditioned imagination, written out as explicit reasoning trajectories, so that the LVLM learns when and where to imagine rather than only how a view changes under a given motion. Test-time simulators and training-time internalization are complementary and could in principle be combined (Tab.~\ref{tab:comp-external}).
\section{\MethodName}

\MethodName{} (Fig.~\ref{fig:method}) uses an allocentric novel-view generator (NVG) $G$ as a training-time spatial simulator that converts answer-only spatial QA data into two complementary forms of supervision. First, $G$ is fine-tuned to render the observation from a queried allocentric viewpoint, and we introduce a Geometry Perturbation (GeoP) metric to assess the geometric consistency of its outputs (Sec.~\ref{sec:allogen}). We then synthesize explicit ASM trajectories for SFT, in which a pretrained LVLM plans task-relevant viewpoints, $G$ renders the corresponding views, and the rendered views are then replaced by textual captions so that the model learns to imagine, rather than receive, the missing observations (Sec.~\ref{sec:asm-sft}). Finally, we construct automatically verifiable perspective-taking questions for reinforcement learning (RL), which expose the model to observations from diverse viewpoints and reinforce the association between a viewpoint description and its observation (Sec.~\ref{sec:asm-rl}). The trajectories supervise the full plan-imagine-reason procedure, while the RL tasks strengthen viewpoint-conditioned imagination. 

\subsection{Allocentric Novel-View Generator}
\label{sec:allogen}
\begin{figure}[t]
\centering
\includegraphics[width=1.0\textwidth]{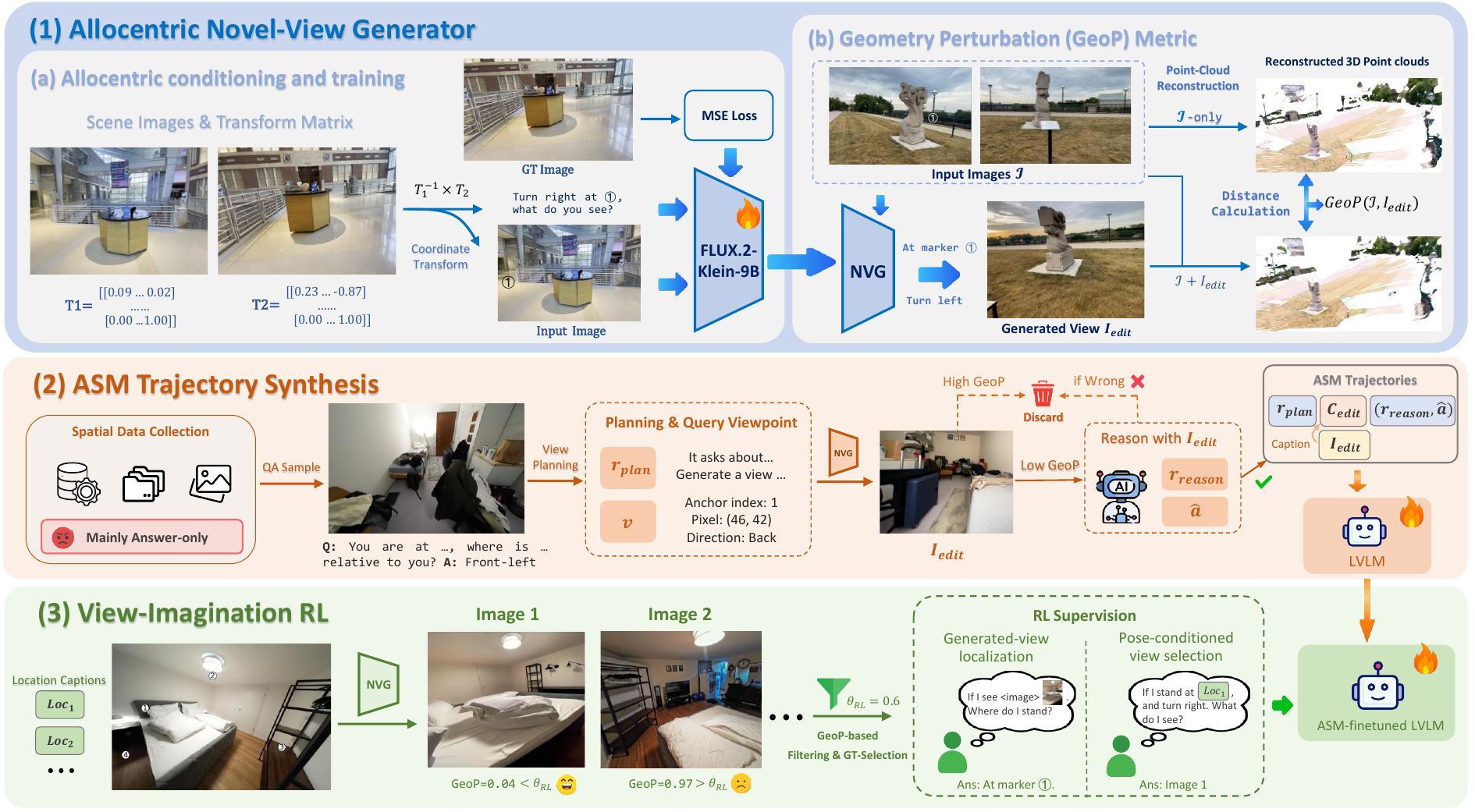}
\vspace{-12pt}
\caption{\textbf{Overview of \MethodName{}.} \textbf{(1)} An allocentric novel-view generator (NVG, $G$) renders the scene from a queried viewpoint, and GeoP measures how much a rendered view perturbs the 3D reconstruction of the inputs. \textbf{(2)} An LVLM plans viewpoints, $G$ renders them, and filtered trajectories are converted to captions for SFT. \textbf{(3)} GeoP-filtered candidate views form verifiable perspective-taking questions for RL.}
\label{fig:method}
\end{figure}

\textbf{Allocentric conditioning.} We obtain \MethodName-NVG (denoted as $G$) by fine-tuning a pretrained image-editing model (FLUX.2-klein-9B~\citep{blackforestlabs2025flux2} in our implementation) to render the observed scene from a queried viewpoint (Fig.~\ref{fig:method}(1a)). Most existing world models act as egocentric simulators: a viewpoint change is expressed as a sequence of ego-motion commands with quantified translations and rotations. In contrast, $G$ is conditioned directly on the target viewpoint, so that a task-relevant perspective can be queried in a single step without specifying the camera motion that leads to it.

Given a set of images $\mathcal{I}=\{I_1, \ldots, I_n\}$ of the same scene, we define an \textbf{allocentric viewpoint} as $v=(i, x, y, d)$, where $(x, y)$ is a pixel coordinate on the anchor image $I_i$ that specifies the target observation location and $d$ is the viewing direction at that location. The anchor index $i$ is required because a pixel coordinate is only meaningful relative to one image, and providing the full set $\mathcal{I}$ gives $G$ additional scene context beyond the anchor view. Given $\mathcal{I}$ and $v$, $G$ renders the observation from the queried viewpoint:
\begin{equation}
I_{\mathrm{edit}}=G(\mathcal{I}, v), \quad v=(i, x, y, d).
\label{eq:nvg}
\end{equation}

Each training sample consists of the image set $\mathcal{I}$, an allocentric viewpoint $v=(i, x, y, d)$, and the ground-truth observation $I_{\mathrm{gt}}$ from that viewpoint.
In practice, $(x, y)$ is provided to $G$ as a visual marker drawn on $I_i$, and $d$ as a textual prompt. 
We build such samples from public scene datasets that provide multiple posed observations of the same scene. For a pair of frames from one scene, we take one frame as the anchor $I_i$ and the other as $I_{\mathrm{gt}}$; the relative camera pose between them determines where the camera of $I_{\mathrm{gt}}$ lies in $I_i$ and which direction it faces, which yields $(x, y, d)$.

\textbf{Training.} We fine-tune the image-editing model on these samples, supervising the rendered view $I_{\mathrm{edit}}=G(\mathcal{I}, v)$ with the ground-truth observation $I_{\mathrm{gt}}$. This supervision asks $G$ to preserve the scene content of $\mathcal{I}$ while changing the perspective as specified by $v$; whether the rendered views are geometrically consistent with the scene is assessed with the metric introduced next. Details of the data construction are given in the Appendix~\ref{Appendix:edit140k}.

\textbf{Geometry Perturbation (GeoP) metric.} Standard novel-view synthesis metrics such as PSNR, SSIM, and LPIPS measure image-level similarity to a reference view and therefore do not capture whether a rendered view is geometrically consistent with the scene under large viewpoint changes. 

To assess the geometric plausibility of views rendered by $G$, we introduce GeoP, a model-based metric that measures how much a rendered view perturbs the 3D geometry inferred from the original observations (Fig.~\ref{fig:method}(1b)). Given the image set $\mathcal{I}$ of a scene, we first reconstruct the scene with a pretrained feed-forward 3D reconstruction model $F$, obtaining a point cloud $\mathcal{M}_{\mathrm{orig}}=F(\mathcal{I})$ from the original observations alone.
We then add the rendered view $I_{\mathrm{edit}}=G(\mathcal{I}, v)$ to the observations and run the same reconstruction:
\begin{equation}
\mathcal{I}^{+}=\mathcal{I}\cup\{I_{\mathrm{edit}}\},\quad \mathcal{M}_{\mathrm{aug}}=F(\mathcal{I}^{+}).
\label{eq:geop-aug}
\end{equation}

Both point clouds are produced by $F$ in the same world coordinate system from its predicted depth maps, camera intrinsics, and camera poses, so they can be compared directly. We define $\mathrm{GeoP}(\mathcal{I}, I_{\mathrm{edit}})$ as the Chamfer distance between $\mathcal{M}_{\mathrm{orig}}$ and $\mathcal{M}_{\mathrm{aug}}$, \textit{i.e.}, the average squared nearest-neighbor distance between the two point sets in both directions:
\begin{align}
\mathrm{GeoP}(\mathcal{I}, I_{\mathrm{edit}}) &= \frac{1}{|\mathcal{M}_{\mathrm{orig}}|}
\sum_{\mathbf{p}\in\mathcal{M}_{\mathrm{orig}}}
\min_{\mathbf{q}\in\mathcal{M}_{\mathrm{aug}}}
\|\mathbf{p}-\mathbf{q}\|_2^2 \nonumber \\
&+ \frac{1}{|\mathcal{M}_{\mathrm{aug}}|}
\sum_{\mathbf{q}\in\mathcal{M}_{\mathrm{aug}}}
\min_{\mathbf{p}\in\mathcal{M}_{\mathrm{orig}}}
\|\mathbf{q}-\mathbf{p}\|_2^2.
\label{eq:geop}
\end{align}

A low GeoP means that adding $I_{\mathrm{edit}}$ changes the reconstructed geometry little, \textit{i.e.}, $F$ can explain the rendered view as another observation of the same scene. A high GeoP means that the rendered view forces $F$ to alter its scene-level geometry, which we take as a sign of geometric inconsistency. GeoP is therefore a proxy that depends on $F$ rather than a ground-truth measure; we use it to filter rendered views with thresholds (Sec.~\ref{sec:asm-sft} and Sec.~\ref{sec:asm-rl}) and to compare generators in our experiments.



\subsection{ASM Trajectory Synthesis}
\label{sec:asm-sft}
\textbf{Motivation.} Existing spatial datasets offer large-scale supervision, but most contain only question-answer (QA) pairs without intermediate reasoning. 
Such answer-only supervision does not teach a model how to perform ASM on spatial tasks that require reasoning beyond the given observations.
Meanwhile, current LVLMs often select uninformative viewpoints and imagine novel views unreliably (Sec.~\ref{sec:preliminary}). Directly prompting an LVLM to write ASM trajectories therefore risks inaccurate or hallucinated intermediate steps, which would yield misleading supervision.

\textbf{Trajectory synthesis.} We therefore synthesize ASM trajectories automatically, using $G$ as an external spatial simulator during data construction only (Fig.~\ref{fig:method}(2)). From large-scale open-source spatial datasets, we select subtasks that require active perspective exploration beyond direct recognition, \textit{e.g.}, reasoning about camera-object relationships and spatial configurations, associating multiple input views, and inferring what is seen from a viewpoint other than the original ones.

Each QA sample from the selected subtasks is a triplet $(q, \mathcal{I}, a)$, where $q$ is the question, $\mathcal{I}$ is the set of input images, and $a$ is the ground-truth answer from the source dataset. For each sample that $M$ fails to answer from $\mathcal{I}$ alone (see \textbf{Filtering} below), we prompt $M$ to produce a view plan with two parts: a reasoning trace $r_{\mathrm{plan}}$ that infers which viewpoints would help answer $q$, and $K$ allocentric viewpoints $\{v_k\}_{k=1}^{K}$, each of which can be used directly as the condition of $G$:
\begin{equation}
(r_{\mathrm{plan}}, \{v_k\}_{k=1}^{K}) = M(q, \mathcal{I}), \quad v_k=(i_k, x_k, y_k, d_k).
\label{eq:plan}
\end{equation}

The planned viewpoints $\{v_k\}$ specify where and in which direction the model should observe the scene to answer $q$. $G$ then renders a view for each of them:
\begin{equation}
I_{\mathrm{edit},k}=G(\mathcal{I}, v_k),\quad k=1,\ldots,K.
\label{eq:render}
\end{equation}

The rendered views are then returned to $M$ together with a textual description $t_{\mathrm{des}}$ of the planned viewpoints, and $M$ answers the question again, producing a reasoning trace $r_{\mathrm{reason}}$ and an answer $\hat{a}$:
\begin{equation}
(r_{\mathrm{reason}}, \hat{a})=M\big(q, t_{\mathrm{des}}, \mathcal{I}, \{I_{\mathrm{edit},k}\}_{k=1}^{K}\big).
\label{eq:reason}
\end{equation}

\textbf{Filtering.} We retain a trajectory only if the rendered views are both useful and geometrically consistent: $M$ must answer incorrectly from $\mathcal{I}$ alone but correctly after seeing the rendered views, and every rendered view must have a GeoP below a threshold $\theta$. Let $\hat{a}_{0}$ denote the answer of $M$ from $\mathcal{I}$ alone. It is checked first, so that questions $M$ already answers correctly are discarded before any viewpoint is planned or rendered (Fig.~\ref{fig:method}(2)). The set of retained image-form trajectories for a question $(q, \mathcal{I})$ is
\begin{equation}
\mathcal{T}_{\mathrm{img}}=\big\{(r_{\mathrm{plan}},\{I_{\mathrm{edit},k}\}_{k=1}^{K},r_{\mathrm{reason}},\hat{a}) \,\big|\,
\hat{a}_{0}\neq a,\ \hat{a}=a,\ \max_{k}\mathrm{GeoP}(\mathcal{I}, I_{\mathrm{edit},k})<\theta\big\},
\label{eq:filter}
\end{equation}
where $\theta$ is a pre-defined threshold. The first two conditions select cases in which the rendered views change the outcome, so that the supervision reflects useful perspective exploration rather than redundant visual input; the third removes geometrically inconsistent views.

\textbf{Caption substitution.} To internalize view imagination rather than view consumption, we replace each rendered view in a retained trajectory with a textual caption. A captioning model $M_{\mathrm{cap}}$ describes $I_{\mathrm{edit},k}$ as $C_{\mathrm{edit},k}=M_{\mathrm{cap}}(I_{\mathrm{edit},k})$ and is prompted to focus on the 2D layout and relative positions of objects in the view. The final ASM trajectories for a question $(q, \mathcal{I})$ are
\begin{equation}
\mathcal{T}_{\mathrm{ASM}}=\big\{(r_{\mathrm{plan}}, \{C_{\mathrm{edit},k}\}_{k=1}^{K}, r_{\mathrm{reason}}, \hat{a}) \,\big|\, (r_{\mathrm{plan}}, \{I_{\mathrm{edit},k}\}_{k=1}^{K}, r_{\mathrm{reason}}, \hat{a})\in\mathcal{T}_{\mathrm{img}}\big\}.
\label{eq:traj}
\end{equation}

Each trajectory thus supervises the model to plan informative viewpoints for the task, to imagine the observation and spatial layout from each planned viewpoint in text, and to reason with the imagined views. Because the rendered images never appear in the supervision, the model learns to carry out this procedure from the original observations alone and does not call $G$ at inference time. This supervision targets multi-step spatial tasks that require information beyond the original observations; its effect is examined in our experiments. 

\subsection{View-Imagination Reinforcement Learning}
\label{sec:asm-rl}

\textbf{Motivation.} To further strengthen the ability of LVLMs to imagine a scene from a designated allocentric viewpoint, we construct automatically verifiable multiple-choice questions and train the SFT model from Sec.~\ref{sec:asm-sft} on them with GRPO~\citep{shao2024deepseekmath} (Fig.~\ref{fig:method}(3)). These questions train the model to associate an allocentric viewpoint query with its observation, without manually annotated reasoning traces. 

\textbf{Candidate views.} For an image set $\mathcal{I}$ of a scene, we randomly sample four allocentric viewpoints $\{v_k=(1, x_k, y_k, d_k)\}_{k=1}^{4}$ anchored on $I_1$ and render one candidate view for each with $G$:
\begin{equation}
\mathcal{V}=\{I_{\mathrm{edit},1},\ldots,I_{\mathrm{edit},4}\},\quad I_{\mathrm{edit},k}=G(\mathcal{I}, v_k).
\label{eq:cand}
\end{equation}
From each candidate set $\mathcal{V}$ we build two complementary multiple-choice tasks. \textit{1) Generated-view localization}. We overlay four numbered markers on the anchor image $I_1$ at the sampled locations $\{(x_k, y_k)\}$ to obtain $I_{\mathrm{mark}}$, and pick one candidate $I_{\mathrm{edit},j}$. Given $(I_{\mathrm{mark}}, \mathcal{I}, I_{\mathrm{edit},j})$, the model must determine at which numbered marker the candidate view was captured and the answer is $j$.
\textit{2) Pose-conditioned view selection}. This task is the inverse: given a textual target pose, the model must pick the matching candidate. For a chosen index $j$ (selected as described below), we use $M_{\mathrm{cap}}$ to describe the location $(x_j, y_j)$ on $I_1$ and combine this description with the viewing direction $d_j$ into a target-pose specification $t_j$. Given $t_j$, $\mathcal{I}$, and the four candidates $\mathcal{V}$, the model must select the candidate that depicts the scene from the described location and direction and the answer is $j$.

\textbf{Answer selection.} The ground-truth candidate $I_{\mathrm{edit},j}$ must be geometrically consistent with $\mathcal{I}$ for the reward to be meaningful, so we select $j$ for each candidate set $\mathcal{V}$ with GeoP:
\begin{equation}
\mathcal{K}=\{k \mid \mathrm{GeoP}(\mathcal{I}, I_{\mathrm{edit},k})<\theta_{\mathrm{RL}}\},\quad
j=\begin{cases}
\text{any } k\in\mathcal{K}, & \mathcal{K}\neq\emptyset,\\
\text{discard the sample}, & \mathcal{K}=\emptyset,
\end{cases}
\label{eq:rl-select}
\end{equation}
where $\theta_{\mathrm{RL}}$ is a pre-defined threshold. The two tasks provide complementary supervision: \textit{generated-view localization} learns the inverse mapping from an observation to its allocentric viewpoint, whereas \textit{pose-conditioned view selection} learns the forward mapping from a textual target pose to its expected observation. Both answers are derived automatically without human annotation, and the two tasks are intended to improve the ability of LVLMs to imagine and reason about scene layout and object configuration from viewpoints other than the observed ones.
\section{Experiments}
\begin{figure}[t]
\centering
\includegraphics[width=1.0\textwidth]{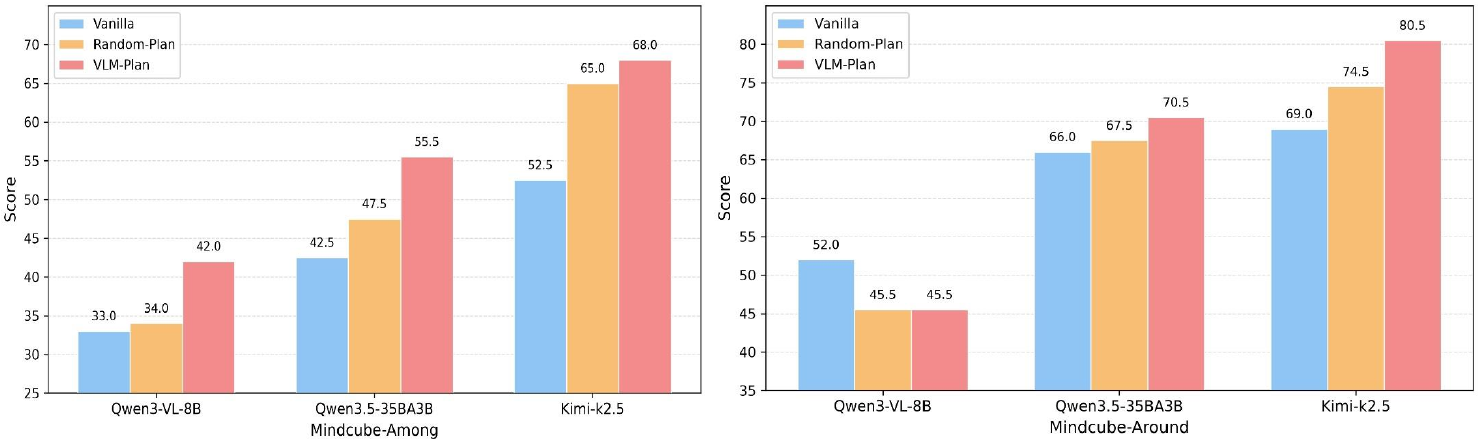}
\vspace{-12pt}
\caption{\textbf{Evaluation of external ASM}. We evaluate vanilla VLM reasoning, and two external ASM strategies with different viewpoint-planning methods. Both planning-based settings invoke \MethodName-NVG and condition the final reasoning on the original and generated observations.}
\label{fig:explicit1}
\end{figure}

\begin{table}[t]
  \caption{\textbf{Performance comparison} on 7 spatial reasoning and 3 general-VQA benchmarks. We report the average accuracy over all benchmarks within each category. The performance of several spatial models are also provided (7B variants on Qwen2.5-VL-7B; 8B variants on Qwen3-VL-8B).}
  \vspace{-12pt}
  \centering
  \setlength{\tabcolsep}{2.4pt}
  \normalsize
  \begin{tabular}{@{}lcccccccc|c@{}}
    \toprule
    \multirow{2}{*}{Models} & \multicolumn{8}{c}{Spatial Reasoning} & \footnotesize{Gnr.-VQA} \\
    \cmidrule(lr){2-9} \cmidrule(l){10-10}
     & Mindcube & MMSI & $\text{Spa457}_\text{6d}$ & Viewspa & 3DSR & EmbSpa & ESI & Avg. & Avg.
      \\
    \midrule
    \rowcolor{gray!10}
    \multicolumn{10}{c}{\textit{Representative Spatial Models}} \\
    SpaceR-7B  & 36.6 & 28.8 & 45.1 & 38.4 & 54.5 & 37.3 & 40.2 & 40.1 & 59.4 \\
    VST-RL-7B  & 39.3 & 35.1 & 43.4 & 47.0 & 60.1 & 72.0 & 40.1 & 48.1 & 62.1  \\
    HoliSpatial-8B & 52.0 & 30.6 & 30.5 & 50.1 & 49.7 & 73.8 & 41.8 & 46.9 & 63.1 \\
    SenseNova-8B & 50.4 & 29.6 & 41.3 & 46.4 & 55.2 & 70.8 & 36.5 & 47.2 & 63.7  \\
    \midrule
    \rowcolor{gray!10}
    \multicolumn{10}{c}{\textit{Baselines \&  Our Model (Qwen3-VL-4B)}}\\
    Qwen3-VL-4B & 33.9 & 28.0 & 37.2 & 43.2 & 55.8 & 72.3 & 36.5 & 43.8 & 63.7 \\
    \MethodName-4B & 39.1 & 29.2 & 42.8 & 46.2 & 56.4 & 76.5 & 40.6 & 47.3 & 63.1 \\
    \rowcolor{blue!6}
    \textit{Improvement} & 
    \textcolor[RGB]{0, 150, 50}{+5.2} & 
    \textcolor[RGB]{0, 150, 50}{+1.2} &
    \textcolor[RGB]{0, 150, 50}{+5.6} &
    \textcolor[RGB]{0, 150, 50}{+3.0} &
    \textcolor[RGB]{0, 150, 50}{+0.6} &
    \textcolor[RGB]{0, 150, 50}{+4.2}& \textcolor[RGB]{0, 150, 50}{+4.1} & \textbf{\textcolor[RGB]{0, 120, 50}{+3.5}}  &
    \textcolor[RGB]{100, 0, 0}{-0.6}
     \\
    \midrule 
    \rowcolor{gray!10}
    \multicolumn{10}{c}{\textit{Baselines \&  Our Model (Qwen3-VL-8B)}}\\

    Qwen3-VL-8B & 40.4 & 27.8 & 34.8 & 45.3 & 57.2 &  79.5 & 38.4 & 46.2 & 64.4\\
    \MethodName-8B & 43.9 & 31.1 & 43.5 & 46.8 & 58.2 & 80.2 & 40.2 & 49.1 & 65.0 \\
    
    \rowcolor{blue!6}
    \textit{Improvement} & 
    \textcolor[RGB]{0, 150, 50}{+3.5} & 
    \textcolor[RGB]{0, 150, 50}{+3.3} &
    \textcolor[RGB]{0, 150, 50}{+8.7} &
    \textcolor[RGB]{0, 150, 50}{+1.5} &
    \textcolor[RGB]{0, 150, 50}{+1.0} &
    \textcolor[RGB]{0, 150, 50}{+0.7} & \textcolor[RGB]{0, 150, 50}{+1.8}& \textbf{\textcolor[RGB]{0, 120, 50}{+2.9}}&
    \textbf{\textcolor[RGB]{0, 120, 50}{+0.6}} \\

\midrule 
    \rowcolor{gray!10}
    \multicolumn{10}{c}{\textit{Baselines \&  Our Model (Qwen3-VL-30BA3B)}}\\

    \small{Qwen3-VL-30BA3B} & 36.7  & 25.2 & 41.9 & 43.9 & 54.5 & 73.4 & 37.2 & 44.7 & 69.4  \\
    \small{\MethodName-30BA3B} & 37.9  & 30.8 & 42.1 & 45.6 & 58.8 & 75.2 &40.3 & 47.2 & 68.9\\
    
    \rowcolor{blue!6}
    \textit{Improvement} & \textcolor[RGB]{0, 150, 50}{+1.2} &
    \textcolor[RGB]{0, 150, 50}{+5.6} & \textcolor[RGB]{0, 150, 50}{+0.2}
    & \textcolor[RGB]{0, 150, 50}{+1.7} & 
    \textcolor[RGB]{0, 150, 50}{+4.3}&  \textcolor[RGB]{0, 150, 50}{+1.8}&
    \textcolor[RGB]{0, 150, 50}{+3.1} &\textbf{\textcolor[RGB]{0, 120, 50}{+2.5}} & 
    \textcolor[RGB]{100, 0, 0}{-0.5} \\
    
    \midrule 
    \rowcolor{gray!10}
    \multicolumn{10}{c}{\textit{Baselines \&  Our Model (Qwen3.5-9B)}}\\
    
    Qwen3.5-9B & 50.8 & 33.5 & 41.4 & 47.3 & 64.4 & 81.4 & 39.1 & 51.1 & 78.1  \\
    \MethodName-9B &  54.7 & 35.2 & 43.5 & 47.9 & 63.5 & 82.3 & 39.2 & 52.3 & 78.5\\
    
    \rowcolor{blue!6}
    \textit{Improvement} & \textcolor[RGB]{0, 150, 50}{+3.9} &
    \textcolor[RGB]{0, 150, 50}{+1.7} & \textcolor[RGB]{0, 150, 50}{+2.1}
    & \textcolor[RGB]{0, 150, 50}{+0.6} & 
    \textcolor[RGB]{100, 0, 0}{-0.9}& \textcolor[RGB]{0, 150, 50}{+0.9} & \textcolor[RGB]{0, 150, 50}{+0.1} & \textbf{\textcolor[RGB]{0, 120, 50}{+1.2}} & 
    \textbf{\textcolor[RGB]{0, 120, 50}{+0.4}} \\
    \bottomrule
  \end{tabular}
  \vspace{-8pt}
  \label{tab:main}
\end{table}

\textbf{Dataset construction.} To train the NVG in Sec.~\ref{sec:allogen}, we leverage \textit{DL3DV}~\citep{ling2024dl3dv}, a public dataset which provides 10k diverse scenes captured from multiple viewpoints together with the corresponding camera matrices, to construct our training set \textbf{SpatialASM-edit140k}, containing both single-image and multi-image inputs. To synthesize ASM trajectories introduced in Sec.~\ref{sec:asm-sft}, we source various challenging spatial-QA samples from \textit{Holi-spatial}, \textit{VST-500k}, and \textit{Mindcube-train}~\citep{gao2026holi, yang2026visual, wang2025spatial} and generate a dataset with 128k reasoning trajectories, denoted \textbf{SpatialASM-128k}. In practice, we use the subset \textbf{SpatialASM-66k} constructed exclusively from \textit{Holispatial} for tuning LVLMs. For RL (Sec.~\ref{sec:asm-rl}), we only use the original 2D observations collected from \textit{HoliSpatial} and \textit{DL3DV}, and automatically construct \textbf{SpatialASM-RL-58k} by employing the NVG. More details are provided in Appendix~\ref{Appendix:dataset}.

\textbf{Training setup.} To internalize ASM capabilities into LVLMs, we adopt the two-stage post-training strategy: first SFT then GRPO. During SFT, we train four models from both the Qwen3-VL and the Qwen3.5 series~\citep{bai2025qwen3, qwen3.5} on \MethodName-66k for 10 epochs with the learning rate of 5e-6. For GRPO, we use a global batch size of 128 and a learning rate of 1e-6, and train for 300 steps. We fully fine-tune Qwen3-VL-4B and Qwen3-VL-8B, while applying LoRA (rank=32) to Qwen3-VL-30B-A3B and Qwen3.5-9B. Training details are provided in Appendix~\ref{Appendix:training}.

\textbf{Evaluation setup.} We evaluate all models on seven Spatial Reasoning (SR) benchmarks~\citep{wang2025spatial,yang2026mmsi,wang2025spatial457,li2026viewspatial,ma20253dsrbench,du2024embspatial,hong2026esi} and three general-VQA benchmarks~\citep{yue2024mmmu,chen2024we,xai2024realworldqa}. All benchmarks except ESI-Bench are evaluated using VLMEvalKit~\citep{duan2024vlmevalkit}. For ESI-Bench, we adopt the passive-single setting. We ask the models to produce reasoning content first on the SR tasks, and to answer directly on the general-VQA tasks. Benchmark and evaluation details are provided in Appendix~\ref{Appendix:benchmark} and Appendix~\ref{Appendix:evaluation}.

\subsection{Preliminary Analysis of Explicit ASM}
\label{sec:preliminary}

Before internalizing ASM into LVLMs, we first wonder whether ASM is an effective reasoning paradigm and whether existing LVLMs lack this capability. Like previous studies in Fig.~\ref{fig:comparison}(b), we conduct a preliminary experiment on the external ASM settings, where LVLMs first plan question-relevant viewpoints, invoke the NVG to synthesize observations from these viewpoints, and then reason jointly over the original and generated images. We sample 200 questions from each of the \textit{Among} and \textit{Around} subtasks of MindCube and report the results of three reasoning methods: (i) Vanilla: LVLMs directly reason over the original observations; (ii) Random-Plan: randomly selected target viewpoints are passed to \MethodName-NVG, and (iii) VLM-Plan: LVLMs are explicitly prompted to plan question-relevant viewpoints before invoking \MethodName-NVG.

By comparing \textit{Vanilla} with the other two settings in Fig.~\ref{fig:explicit1}, the planning-based external ASM paradigm generally improves accuracy over vanilla reasoning, which demonstrates that viewpoint planning and imagination constitute a generally effective paradigm for complex spatial reasoning. Moreover, LVLM-based planning consistently matches or outperforms random planning, indicating that existing LVLMs already possess a certain ability to identify informative viewpoints when explicitly prompted. Nevertheless, this ability is not spontaneously activated during direct LVLM reasoning. In other words, existing LVLMs are able to execute view-planning only under an explicitly-prompted inference pipeline. ASM has not yet become an intrinsic reasoning capability that they can autonomously invoke when needed for certain complicated spatial tasks. 

Furthermore, Qwen3-VL-8B drops under both random and LVLM-based planning, suggesting that relatively small models still lag behind in both robust planning for diverse spatial tasks and long-context multimodal reasoning. It further motivates internalizing ASM through training so that models can learn when and how to employ this reasoning paradigm more reliably.

\subsection{Main Results}

\begin{table}[t]
  \caption{\textbf{Accuracy and average inference time of different methods on benchmark subtasks.} (i) Vanilla: the LVLM reasons directly from the input observations alone, and the reasoning duration is reported; (ii) + External ASM: Similar to Sec.~\ref{sec:preliminary}, the LVLM externally invokes \MethodName-NVG to help reasoning. The time to synthesize novel-view observations is reported.}
  \vspace{-11pt}
  \centering
  \setlength{\tabcolsep}{2.5pt}
  \normalsize
  \begin{tabular}{@{}l|cc|cc|cc|cc@{}}
    \toprule
    \multirow{3}{*}{Subtask} & \multicolumn{4}{c}{Qwen3-VL-8B} & \multicolumn{4}{c}{\MethodName-8B} \\
    \cmidrule(lr){2-5} \cmidrule(lr){6-9}
    & \multicolumn{2}{c}{Vanilla} & \multicolumn{2}{c}{+ External ASM} & \multicolumn{2}{c}{Vanilla} & \multicolumn{2}{c}{+ External ASM}  \\
    \cmidrule(lr){2-3} \cmidrule(lr){4-5} \cmidrule(l){6-7} \cmidrule(l){8-9}
     & Accuracy & T(s) & Accuracy & $\text{T}_\text{NVG}$(s) & Accuracy & T(s) & Accuracy & $\text{T}_\text{NVG}$(s)
      \\
    \midrule
$\text{Mindcube}_{\text{Among}}$ &   33.0 & 12.6 & 42.0 & 57.6  & 39.5 & 18.6 & \textbf{45.0} & 56.8 \\   $\text{Mindcube}_{\text{Around}}$ & 52.0 & 7.5 & 45.5 & 38.1  & 58.0 & 7.2 & \textbf{61.5} & 46.8  \\
    \bottomrule
  \end{tabular}
  \vspace{-8pt}
  \label{tab:comp-external}
\end{table}

\textbf{Overall SR performance gains.} As shown in Table~\ref{tab:main}, \MethodName{} improves spatial reasoning on both Qwen3-VL and Qwen3.5 architectures. The Qwen3-VL models of three different scales, including dense models and an mix-of-experts (MoE) model, yield consistent performance gains on all seven SR benchmarks after applying our approach. In particular, the 4B model achieves an average improvement of \textbf{+3.5\%}, with notable gains on Mindcube (+5.2\%) and Spatial457 (+5.6\%). In contrast, prior spatial models exhibit uneven performance across benchmarks. For example, Sensenova-8B improves significantly on Mindcube (+10.0\%), but exhibits drops by 8.7\% on EmbSpa. This suggests that their gains may depend more on benchmark-specific task formats, whereas \MethodName{} elicits a more universal and transferable spatial capability.

\textbf{Analysis of spatial capabilities.} The largest gains generally occur on more challenging benchmarks (e.g., MindCube, MMSI, and $\text{Spatial457}_\text{6d}$) which require the comprehensive use of multiple reasoning capabilities, such as spatial mental modeling, cross-view correspondence, complex 6D reasoning, etc. These tasks benefit from the learned ability to plan informative viewpoints and imagine the corresponding observations. SpatialASM also yields modest gains on relatively perception-centric SR benchmarks (e.g., 3DSR, Embspa), indicating that the knowledge learned from ASM reasoning trajectories and the RL tasks improves the general understanding of the 3D space.

\textbf{General Visual Capabilities.} \MethodName{} preserves the general visual capabilities of LVLMs, with changes in the average general-VQA accuracy remaining within 0.6\% across all four backbones. Importantly, \MethodName-8B outperforms the two 8B counterparts in \textit{representative spatial models}, where HoliSpatial experiences an average degredation of 1.3\% on all three benchmarks after SFT. We attribute this result partly to view-imagination RL, which exposes LVLMs to diverse visual observations and high-quality, geometry-filtered NVG outputs. Consequently, the model learns from rich visual supervision rather than merely memorizing the reasoning patterns of the SFT trajectories. The models' performance on each benchmark is shown in Appendix~\ref{Appendix:general}.

\textbf{Comparison with external ASM.} As shown in Table~\ref{tab:comp-external}, external ASM is not always beneficial for the baseline model. Introducing redundant visual evidence or unnecessary view-planning CoT may degrade performance (52.0\% to 45.5\% on \textit{Around}). In contrast, after internalization, \MethodName-8B improves under vanilla reasoning on both subtasks without noticeable increase on the inference duration. \textbf{Most importantly}, combining \MethodName-8B with external ASM yields the highest accuracy on both Among (45.0\%) and Around (61.5\%). \MethodName{} strengthens the model’s intrinsic ability to plan viewpoints and reason about spatial evidence, while external NVG provides additional informative views. Employing them together achieves higher accuracy than either approach alone, suggesting that internalization and external tool-use provide complementary gains.

\subsection{Ablation Studies}

\textbf{Comparison of novel-view generators}. Following the experimental design in Sec.~\ref{sec:preliminary}, we execute the external ASM reasoning paradigm with different novel-view generators and evaluate on the same Mindcube subsets as in Sec.~\ref{sec:preliminary}. The accuracy as well as the median GeoP value of all generated images are reported. As shown in Table~\ref{tab:comp-NVG}, compared with other world models, \MethodName-NVG achieves \textbf{the highest accuracy} for both backbone LVLMs on both Mindcube subsets. 

Interestingly, \MethodName-NVG does not achieve the best GeoP as SVC obtains lower GeoP values for both LVLMs, but it consistently yields substantially higher reasoning accuracy. This observation suggests that \textbf{geometric consistency alone} does not fully determine the utility of generated views for downstream spatial reasoning. We attribute this to our allocentric form of viewpoint representation.  Rather than describing the target viewpoint through a sequence of quantified egocentric motions, LVLMs can designate the viewpoint through a single query, and the corresponding location of the viewpoint is explicitly marked in the original input image.  It makes the spatial correspondence between the original scene and the generated observation easier for LVLMs to understand during final multi-image reasoning. Therefore, our allocentric-view generation provides not only reliable novel views, but also auxiliary evidence that is more directly aligned with the reasoning objective.

\textbf{Training and Data recipe}. Table~\ref{tab:abl} compares the independent effect of each training stage and evaluates the filtering strategy for data preparation. All of the four single-stage variants are initialized from the  baseline LVLM. For both SFT and RL, applying GeoP-based filtering improves the average performance on the four SR benchmarks, indicating that GeoP effectively removes geometrically distorted novel views and prevents unreliable visual evidence from being incorporated into ASM reasoning trajectories or RL supervision.

\begin{table}[t]
  \centering
  \begin{minipage}[t]{0.45\linewidth}
    \centering
    \footnotesize
    \caption{Comparison of different novel-view generators under external ASM paradigm.}
    \vspace{-3.5mm}
    \setlength{\tabcolsep}{2.0pt}
    \label{fig: Analysis about number of QA numbers.}
    \begin{tabular}{l|ccc}
      \toprule
      Methods & \small{$\text{Acc}_\text{among}\uparrow$} &  \small{$\text{Acc}_\text{around}\uparrow$} & \small{GeoP}$\downarrow$\\
      \midrule
    Qwen3.5-35BA3B & 42.5 & 66.0  & / \\
    \midrule      
    +GPT-Image2 & 42.0 & 69.0 & 0.141 \\
      +SpatialEdit & 46.0 & 68.0 & \underline{0.110} \\
      +SVC  & \underline{48.5} & \underline{68.5} & \textbf{0.087}  \\
      \small{+\MethodName-NVG} & \textbf{55.5} & \textbf{70.5} & 0.118  \\
       \midrule
     Kimi-k2.5 & 52.5 & 69.0 & / \\
      \midrule
      +GPT-Image2 & 52.0 & 71.0 & 0.157 \\
      +SpatialEdit & \underline{52.5} & 69.0 & 0.138 \\
    +SVC  & \underline{52.5} & \underline{74.0} & \textbf{0.072} \\
      \small{+\MethodName-NVG} & \textbf{68.0} & \textbf{80.5} & \underline{0.099}\\
      \bottomrule
    \end{tabular}
    \label{tab:comp-NVG}
  \end{minipage}
  \hfill
  \begin{minipage}[t]{0.48\linewidth}
    \centering
    \footnotesize
    \caption{Ablations on training strategies and GeoP-based data filtering strategies.}
\vspace{-3.5mm}
    \label{fig: Ablations about N.}
    \renewcommand{\arraystretch}{0.95}
    \setlength{\tabcolsep}{1.0pt}
    \begin{tabular}{l|ccccc}
      \toprule
      \scalebox{0.95}{Methods} & \scalebox{0.95}{Mindcube} & \scalebox{0.95}{EmbSpa} & ViewSpa & \scalebox{0.95}{ESI} & \scalebox{0.95}{Avg.} \\
      \midrule
      \scalebox{0.95}{Qwen3-VL-4B} & 33.9 & 72.3 & 43.2 & 36.5 & 46.5  \\
      \midrule
      \rowcolor{blue!6}
      \multicolumn{6}{c}{SFT with ASM trajectories}\\ 
      \midrule
      w/o filter & 36.4 & 75.0 & 45.3 & 36.8 & 48.4 \\
      with filter & \underline{38.7} & 74.6 & \textbf{46.2} & \underline{40.2} & \underline{49.9}  \\
      \midrule
      \rowcolor{blue!6}
      \multicolumn{6}{c}{View-imagination RL}\\ 
      \midrule
      w/o filter & 34.2 & 75.7 & 43.8 & 39.2 & 48.2 \\
      with filter & 35.1 & \underline{76.3} & 43.6 & 40.1 & 48.8  \\
      \midrule
      \rowcolor{blue!6}
      \multicolumn{6}{c}{SFT + RL}\\ 
      \midrule
      \scalebox{0.95}{\MethodName-4B} & \textbf{39.1} & \textbf{76.5} & \textbf{46.2} & \textbf{40.6} & \textbf{50.6} \\
      \bottomrule
    \end{tabular}
    \label{tab:abl}
  \end{minipage}
\vspace{-5mm}
\end{table}

The two training stages also exhibit \textbf{distinct but complementary} effects. SFT with \MethodName-66k (with filter) produces strong gains on tasks requiring \textbf{explicit viewpoint transformation} and multi-step spatial reasoning, such as MindCube (+4.8\%) and ViewSpa (+3.0\%), which verifies the high quality of our ASM trajectories and the reliability of the proposed CoT synthesis engine from answer-only spatial QAs. In contrast, view-imagination RL is especially effective on the more \textbf{perception-oriented} EmbSpa benchmark and outperforms the setting of applying SFT alone. Combining SFT and RL achieves the best overall accuracy, suggesting a clear synergy where SFT primarily teaches the structured ASM reasoning pattern while RL further strengthens viewpoint-conditioned spatial perception and imagination. Some detailed design is given in Appendix~\ref{Appendix:ablation}.
\section{Conclusion}
In this work, we identify Active Spatial Mentalization (ASM) as a key bottleneck in LVLM spatial reasoning. To address this issue, we propose SpatialASM, a novel post-training framework that internalizes ASM capabilities into LVLMs through training-time spatial simulation. Experiments across seven spatial reasoning benchmarks and four model architectures demonstrate that SpatialASM consistently improves spatial reasoning while preserving general visual capabilities. More importantly, our framework provides a scalable approach to constructing CoT supervision for existing spatial datasets, addressing the common lack of intermediate reasoning processes in open-source spatial data. In future work, we plan to  internalize other forms of tool-assisted reasoning and extend ASM to dynamic real-world scenes, embodied interaction, and longer-horizon 3D reasoning.

\clearpage
\bibliographystyle{plain}
\bibliography{main}


\clearpage
\appendix
\section{More Related Work}
\label{Appendix:related}
\subsection{Proactive Spatial Intelligence}
LVLMs have exhibited outstanding performance on various tasks, such as image and video captioning~\citep{xing2026caprl, yang2026caprl++}, math reasoning~\citep{dabiriaghdam2026vamps, zhang2025booststep}, agent harness~\citep{team2026kimi3, ding2026wildclawbench}, and document understanding~\citep{feng2026docscope, wu2026scidocbench}. In spatial reasoning, with the progress of LVLMs, recent benchmarks are moving beyond static spatial recognition and understanding towards more demanding spatial reasoning tasks where LVLMs still perform  significantly worse than ordinary human beings. For example, Spatial457~\citep{wang2025spatial457} progressively evaluates spatial understanding from object recognition and 2D localization to complex 6D reasoning, revealing a pronounced performance degradation as geometric complexity increases. MindCube~\citep{wang2025spatial} studies whether LVLMs can construct a coherent spatial mental model from only a few observations, which evaluates cognitive mapping, perspective taking, and mental simulation of hypothetical movements. Similarly, MMSI-Bench~\citep{yang2026mmsi} targets multi-view spatial intelligence and identifies scene reconstruction, cross-view correspondence, and perspective-transformation reasoning as major failure modes of current LVLMs. These benchmarks suggest that successful spatial reasoning often requires more than extracting relations from the observations already provided: the model must reason about how the scene would be perceived under alternative unobserved viewpoints.

The recent ESI-Bench~\citep{hong2026esi} delves deeper into the \textbf{perception-action loop} and treat it as a higher level of spatial intelligence. An agent can selectively employ locomotion, and manipulation to uncover information that cannot be resolved from the initial observation alone. By reasoning under the \textit{active multiple} setting, where the models can proactively execute movements and interact with the simulation environments, the performance significantly outperforms that under the \textit{passive single} setting. However, the accuracy of the \textit{active multiple} setting lags far behind the \textit{GT passive} setting, where the best ground-truth passive trajectories are provided. These results suggest that current LVLMs lack intrinsic proactive exploration capability when solving spatial problems, unless they are explicitly prompted to do so. And their trajectory-planning capability is still far from saturated. 

To internalize the thinking pattern of `perception-active loop' into LVLMs, Geothinker~\citep{li2026thinking} proposes active geometry integration, in which semantic representations selectively query task-relevant geometric features. However, such geometry-enhanced approaches~\citep{zheng2026learning,li2026thinking} introduce an additional
pretrained feature branch and architecture-specific fusion modules into the
LVLMs, and require alignment between geometric and semantic
representations, potentially limiting their adaptability across heterogeneous
VLM architectures. Unlike their notion of `active' at the \textbf{feature-retrieval level}, our ASM operates at the \textbf{viewpoint and observation level}. It reasons about \textit{where} useful evidence should be acquired and
\textit{what} should be observed from the corresponding hypothetical viewpoint. And we focus on data-centric pipeline where such capability can be internalized through off-the-shelf post-training strategies without introducing geometry module or making architectural
modification.

\subsection{Spatial data synthesis}
A wide spectrum of previous work lay emphasis on large-scale high-quality data curation pipeline to improve spatial reasoning in LVLMs. A straightforward approach is to construct such supervision via human annotation~\citep{cai2025spatialbot}, but it is hardly affordable to scale. Some other pipelines resort to proprietary/open-source large-language models~\citep{ro2025well, cheng2024spatialrgpt}. However, current models' performance on relatively challenging spatial reasoning tasks, especially in multi-view or video modalities, remains far from satisfactory and may introduce abundant errors to curated QA pairs.

To formulate scalable spatial supervision with high accuracy and quality, various synthesis frameworks emerge with two predominate characteristics: leveraging public scene datasets with explicit geometric
annotations, and extensively employing visual expert models to recover
otherwise unavailable 3D information from ordinary images. \textbf{i)} Numerous approaches (e.g., SpatialRGPT, OpenSpatial) leverage open-source scene datasets (e.g., ScanNet) with diverse annotated scenes as their foundational data~\citep{li2026spatialladder, ouyang2025spacer,liu2026openspatial}. These datasets provide 3D annotations, such as object classification, semantic voxel labels, and 3D segmentations, which can be easily augmented to spatial QA tasks. \textbf{ii)} A complementary direction further adds necessary annotations for constructing spatial supervision through specialized visual models~\citep{chen2024spatialvlm, cheng2024spatialrgpt,deng2026internspatial}. These models have grown relatively mature in the traditional field of computer vision and perform well in tasks such as segmentation~\citep{liu2024grounding}, metric depth estimation~\citep{hu2024metric3d}, object detection~\citep{zhang2024recognize}, and pose estimation~\citep{wang2024dust3r}, thereby lifting 2D imagery to 3D context.

While these pipelines greatly expand the scale of spatial supervision, the
resulting data are largely constrained by the expert models or the information already encoded in
the source public datasets.
Recent works therefore explore richer sources of supervision. For instance, VST~\citep{yang2026visual} broadens the coverage of spatial supervision
across single-image, multi-image, and video scenarios by leveraging a 3D scene simulator to generate novel observations. Spatial-SSRL~\citep{liu2026spatial} bypasses explicit
spatial annotation by constructing five self-supervised pretext tasks which provides naturally verifiable rewards for spatial reinforcement learning.
Holi-Spatial~\citep{gao2026holi} turns raw video streams into
geometrically optimized 3D Gaussian Splatting (3DGS) scenes and automatically derives
multi-level supervision (e.g.,
3D grounding instances, QA pairs).
SenseNova-SI~\citep{cai2026scaling} scales spatial intelligence to a new level, curating millions of spatial training samples by incorporating and filtering QA pairs from an incredibly large variety of datasets.

Despite the increasing scale and diversity of these datasets, most of them only provide \textbf{answer-only}
supervisory signals. Specifically, for those relatively challenging tasks (e.g., perspective-taking allocentric transformation, mental reconstruction), the answers remain centered on \textbf{what} spatial relation or camera motion should be predicted, rather than explicitly demonstrating \textbf{how} a model can approach the correct answer
through proactively acquiring missing evidence during reasoning. 

VST-R~\citep{yang2026visual} constitutes an important step towards reasoning supervision by providing 135K curated sample with explicit
step-by-step reasoning traces. 
However, its synthesis process requires ground-truth 3D bounding boxes in the original scene data to form intermediate Bird's-Eye View (BEV) images. Moreover, its reasoning supervision primarily teaches the model to derive
spatial conclusions from the provided visual evidence, rather than to execute the \textit{perception-action loop} of proactive exploration. It is not yet publicly released as well. In contrast, \MethodName{} grounds an explicit question-conditioned reasoning process of \emph{when} additional viewpoint reasoning is needed,
\emph{where} the model should mentally observe, and \emph{what} evidence
should be expected from that viewpoint, which is more demanding and human-like.
Importantly, \MethodName{} can synthesize ASM reasoning trajectories for existing spatial QA data without requiring any prerequisite ground-truth 3D annotations.

World2VLM~\citep{zhang2026world2vlm} is particularly relevant to our work in
that it also employs a generative world model in its data engine and
distills its spatial imagination into the LVLMs. World2VLM synthesizes future views based on given parameterized camera trajectories to teach LVLMs to associate multi-view observations with
egocentric motion transitions. Though effective in enhancing view-imagination capabilities, World2VLM fails to teach how to initialize a thinking process with imaginary world-model-like motions when encountering spatial problems, nor does it provide any reasoning content in its supervision.
Our objective is different: SpatialASM starts from the \textbf{spatial question itself} and teaches the model to determine which viewpoint would provide useful missing evidence, rather than starting from a predefined
egocentric motion trajectory and learning the corresponding visual outcome. The generated views serve only as temporary scaffolding for synthesizing
\textbf{\MethodName-128k}, whose final supervision contains long-CoT traces which explicitly couples
\textbf{question-conditioned view planning}, \textbf{viewpoint-conditioned imagination}, and
\textbf{reasoning} with the imagined evidence.

\subsection{Tool-Assisted Spatial Reasoning} Recent studies increasingly augment LVLMs with paradigms of external harness to improve spatial intelligence. The tools can be primarily categorized into three groups: specialized visual tools, 3D reconstruction tools, and generative world models.

\textbf{Specialized visual tools} are computer-vision models which provide more reliable low-level or metric spatial evidence, such as pose/depth estimation and segmentation. Multiple studies demonstrate that the accurate specialist perception substantially improves spatial reasoning when LVLMs are specifically optimized for determining which tools to invoke and how to integrate their outputs~\citep{cho2026spatialclaw, dai2026s, zhou2026spatialcli}. Pretrained \textbf{3D reconstruction} models recover point clouds and camera poses from scene  images~\citep{keetha2026mapanything, lin2025depth} so that LVLMs can
actively manipulate the reconstructed scene and inspect spatial configurations
from alternative viewpoints~\citep{zhang2026think3d}.
A third direction employs \textbf{generative world models} to synthesize observations
that are not present in the original visual inputs~\citep{zhu2026thinking, yu2026and, yang2026mindjourney}.  These methods primarily enhance spatial reasoning by augmenting the external evidence, which introduce additional inference overhead due to
external tool execution and LVLM-tool interaction.
Moreover, specialized tool-use training on LVLMs may introduce an additional
trade-off between tool proficiency and the original general capabilities~\citep{chen2025atlas}.

\MethodName{} targets a fundamentally different yet complementary dimension from
the above tool-assisted approaches: strengthening the intrinsic spatial
reasoning process itself. While tool-assisted methods improve the availability and precision of external spatial evidence, \MethodName{} improves the model's ability to reason about what spatial evidence is needed and how it should be used.
The two paradigms can thus be \textbf{naturally combined}, improving spatial intelligence along complementary axes of intrinsic reasoning capability and
external evidence accessibility.

\section{Dataset Details}
\label{Appendix:dataset}
\subsection{\MethodName-edit140k}
\label{Appendix:edit140k}
\subsubsection{Statistics}
\MethodName-edit140k consists of 139,730 samples, all automatically constructed from the large-scale public scene datasets \textit{DL3DV}~\citep{ling2024dl3dv}. As illustrated in Sec.~\ref{sec:allogen}, each sample has the form of $(\mathcal{I}, v, I_\text{gt})$, containing the input images $\mathcal{I}$, the allocentric viewpoint $v$, and expected ground-truth novel-view observation. To enable $G$ to generate novel views from multi-image inputs, we set the length of input images $|\mathcal{I}|$ to different values, and the number of samples of each $|\mathcal{I}|$  is provided in Tab.~\ref{tab:app-edit}.

\begin{table}[h]
  \caption{The samples numbers of each input-image length.}
  \vspace{-5pt}
  \centering
  \setlength{\tabcolsep}{3.5pt}
  \normalsize
  \begin{tabular}{@{}c|cccc|c@{}}
    \toprule
    \textbf{Input Length} & $|\mathcal{I}|=1$ & $|\mathcal{I}|=2$ & $|\mathcal{I}|=3$ & $|\mathcal{I}|=4$ & \textbf{Total}  \\
    \midrule
    \textbf{Size} & 35,026 & 34,835 & 34,846 & 35,023 & 139,730\\
    \bottomrule
  \end{tabular}
  \vspace{-8pt}
  \label{tab:app-edit}
\end{table}

\subsubsection{Construction Method}
Given a 3D scene video in $DL3DV$, we first select a few observations as $\mathcal{I}=\{I_1, ..., I_n\}$, and another frame as $I_\text{gt}$. Critically, $I_\text{gt}$ is excluded from $\mathcal{I}$ and the frame indices of any two images in $\mathcal{I}$ differ by at least 30, ensuring a noticeable viewpoint difference between them.

Since the image-editing model cannot directly accept the viewpoint query $v=(i,x,y,d)$ as its input, so we separate and integrate it into the image and prompt inputs. In practice, we lay a visual marker on the pixel $(x, y)$ of the anchor image $I_i$, and we transform $d$ into a textual prompt. Here we will explain how we derive $(x, y)$ from the metadata provided in $DL3DV$.

Assume that $I_1$ is the anchor image we select, then we need to compute the viewpoint of $I_\text{gt}$ to derive $(x, y)$ in the query based on the coordinate system of $I_1$. We can easily collect their corresponding camera-to-world transformation matrices $T_1$, $T_\text{gt}\in SE(3)$ from the original dataset. $SE(3)$ denotes the set of 3D rigid transformations, each represented by a
$4\times4$ homogeneous matrix that combines a 3D rotation and a translation. The transformation from
camera-gt to camera 1 is:
\begin{equation}
T_{gt\rightarrow1}=T_1^{-1}T_{gt}
=
\begin{bmatrix}
R_{gt\rightarrow1} & \mathbf{t}_{gt\rightarrow1}\\
\mathbf{0}^{\mathsf T} & 1
\end{bmatrix},
\end{equation}
where $R_{gt\rightarrow1}\in\mathbb{R}^{3\times3}$ is the relative rotation of
camera-gt with respect to camera 1, and
$\mathbf{t}_{gt\rightarrow1} = [t_x, t_y, t_z]^T$ is the position of camera-gt in the coordinate system of camera 1. This position is projected onto $I_1$ as:
\begin{equation}
x=f_x\frac{t_x}{t_z}+c_x,
\qquad
y=f_y\frac{t_y}{t_z}+c_y,
\end{equation}
where $(f_x,f_y)$ are the focal lengths and $(c_x,c_y)$ is the principal point, also provided in $DL3DV$ originally. A valid projection requires $t_z>0$ and the resulting pixel $(x,y)$ to lie inside
the image boundary. Finally, we derive the allocentric viewpoint direction $d$ based on the relative rotation around the $y$-axis, which can be computed from
$R_{gt\rightarrow1}$ and expressed as a pitch angle in degrees. And we assign $d$ to the corresponding direction in
$\{\text{forward},\text{left},\text{back},\text{right}\}$.

\subsubsection{Training samples}

\begin{figure}[h]
\centering
\includegraphics[width=1.0\textwidth]{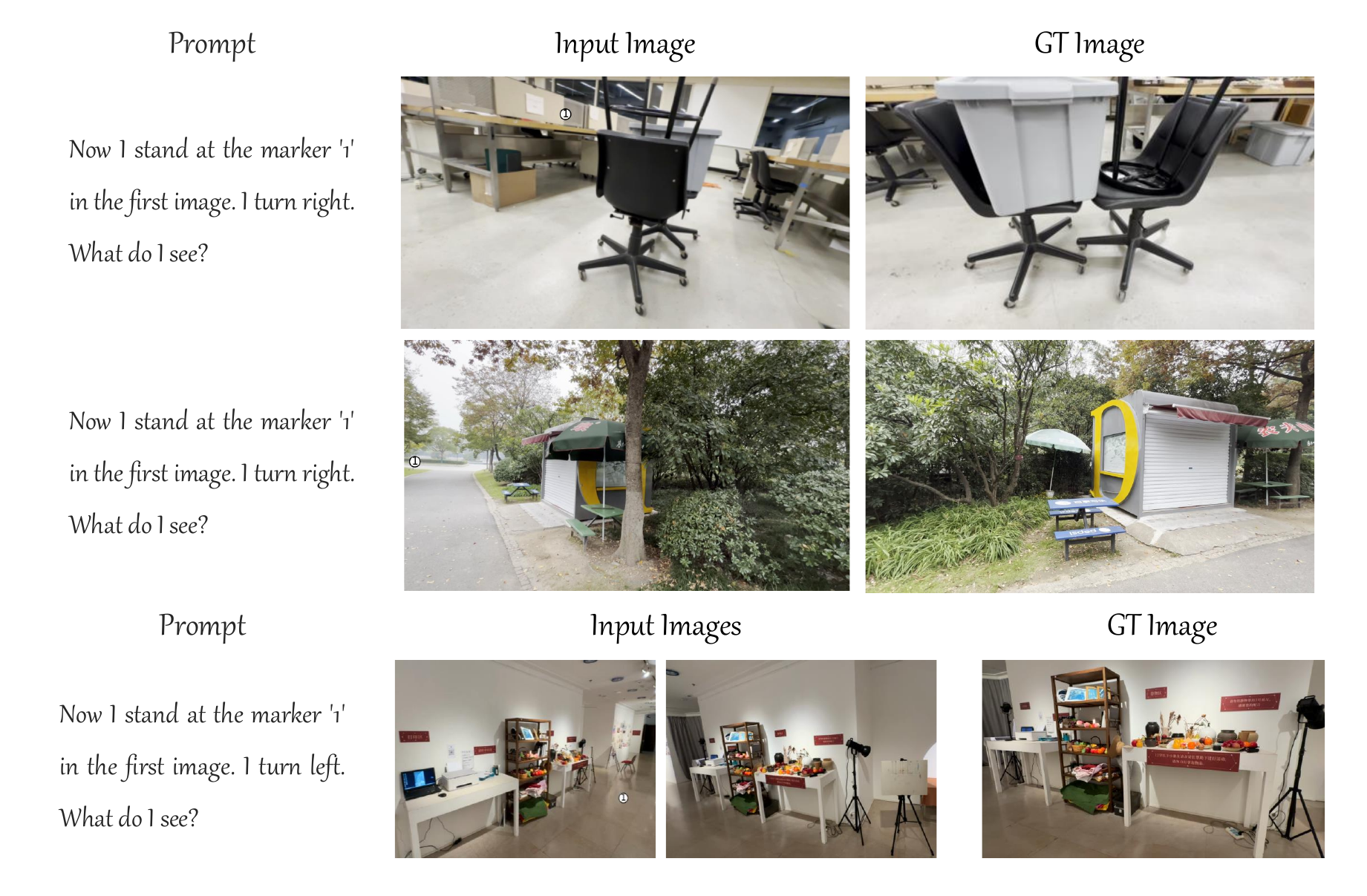}
\vspace{-12pt}
\caption{Examples in \MethodName-edit140k. (Case 1, 2: $|\mathcal{I}|=1$, Case 3: $|\mathcal{I}|=2$)}
\label{fig:app-editcase}
\end{figure}

We provide two examples with single and multiple image inputs in Fig.~\ref{fig:app-editcase}. As in some training samples, when the queried viewpoint reveals regions outside the views of the original visual inputs, we allow and encourage $G$ to imagine objects \textbf{not present} in the input images, provided that these imagined objects are plausible in the real world (e.g., the patio umbrella and the window on the left wall in $I_\text{gt}$ of Case 2).

\subsection{\MethodName-128k}
In Sec.~\ref{sec:asm-sft}, we explain how to synthesize the ASM reasoning trajectories based on triplets $(q, \mathcal{I}, a)$ from existing answer-only spatial datasets. We construct 128k ASM trajectory samples, and use a subset of 66k samples for our SFT training. Our approach provides a scalable and convenient pathway to synthesizing high-quality spatial CoTs as it does not require any other metadata (e.g., BEV image, bounding box). It helps address the scarcity of reasoning content in prior large-scale spatial supervision and elicit active task-conditioned planning and imagination capabilities in LVLMs from scratch. 

\subsubsection{Statistics}
All original spatial-QAs are sourced from Holi-Spatial~\citep{gao2026holi}, VST-500K~\citep{yang2026visual}, and the training set of Mindcube ~\citep{wang2025spatial}. We give the number of samples from each original data source in \MethodName-128k in Tab.~\ref{tab:app-sft128k}. Importantly, to avoid distribution overlap between training data and our final evaluation on benchmarks, \MethodName-66k \textbf{does not contain} any sample synthesized from Mindtube-train.
\begin{table}[h]
  \caption{The number of samples from each data source in \MethodName-128k}
  \vspace{-5pt}
  \centering
  \setlength{\tabcolsep}{3.5pt}
  \normalsize
  \begin{tabular}{@{}c|ccc|c@{}}
    \toprule
    \textbf{Data Source} & Holi-Spatial & VST-500k & Mindcube-train  & \textbf{Total}  \\
    \midrule
    \textbf{Sample No.} & 91,648 & 34,741 & 1,965 & 128,354\\
    \bottomrule
  \end{tabular}
  \vspace{-8pt}
  \label{tab:app-sft128k}
\end{table}

\subsubsection{Synthesis Protocols}
Some detailed implementations for ASM trajectory synthesis are given in this section. In Sec.~\ref{sec:asm-sft}, we have mentioned that we use two pretrained LVLMs, $M$ and $M_{cap}$, to generate certain parts in our CoT trajectories. $M$ is prompted to produce planning trace $r_{plan}$ and the final reasoning trace $r_{reason}$, and $M_{cap}$ is used to transform $\{I_{\mathrm{edit},k}\}_{k=1}^{K}$ into $\{C_{\mathrm{edit},k}\}_{k=1}^{K}$. In practice, we employ \small Qwen3.6-35BA3B~\citep{qwen36_35b_a3b} \normalsize as $M$ and \small Qwen3.6-27B~\citep{qwen36_27b} \normalsize as $M_{cap}$. For filtering, $\theta$ is used to discard samples whose $\{I_{\mathrm{edit},k}\}_{k=1}^{K}$ have high GeoP values. Since we also filter for reasoning correctness in SFT (Equation~\ref{eq:filter}), we adopt a relatively lenient quality threshold for $\{I_{\mathrm{edit},k}\}_{k=1}^{K}$. We set $\theta=2.5$ during our filtering process.

Here we provide the system prompts of \textbf{planning}, \textbf{reasoning}, and \textbf{captioning}. The planning prompt is for producing $(r_{\mathrm{plan}} \{v_k\}_{k=1}^{K})$ in Equation~\ref{eq:plan}; the reasoning prompt is for producing $(r_{reason}, \hat{a})$ in Equation~\ref{eq:reason}; the captioning prompt is for producing $\{C_{\mathrm{edit},k}\}_{k=1}^{K}$ in Equation~\ref{eq:traj}.

\begin{figure}[h]
\centering
\includegraphics[width=0.88\textwidth]{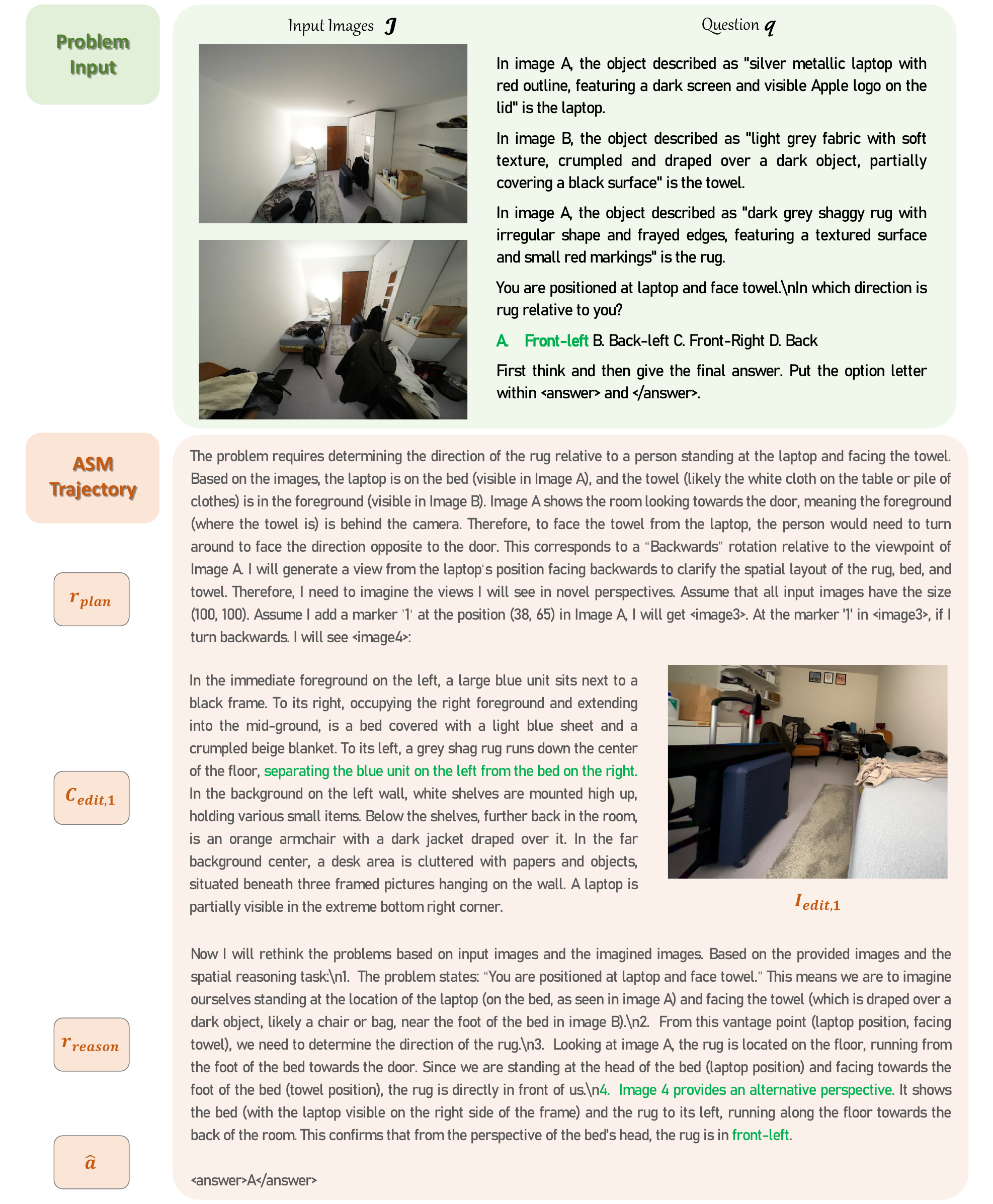}
\vspace{-12pt}
\caption{Example in \MethodName-128k. Note that $I_\text{edit, 1}$ does not exist in the ASM trajectory.}
\label{fig:app-sftcase}
\end{figure}

\begin{itemize}
    \item \textbf{Planning prompt}: I gave you several images and a spatial reasoning question. You can now use an image-editing model to generate images of the scene from new perspectives based on the input images. 
    
    You do not need to answer the spatial reasoning question I gave you, but you need to reason based on the question and give commands to the image-editing model to provide more useful information for answering this question.
    
    For each command, you need to first select a source image from the input images. Then you need to select an area on the image and assign a facing direction, so that the image-editing model will generate the scene viewed from this area and specific facing direction.
    
    Provide your answer in the following format: a list containing Dict(s):
    
    [\{`source\_img': `$<$e.g. 1$>$', `pixel\_x': '$<$e.g. 22$>$', `pixel\_y': `$<$e.g. 63$>$', 'Direction': `e.g. $<$Right$>$'\}, ...]
    
    `source\_img'  is the anchor image number (the first image is numbered 1, and so on). `pixel\_x' and 'pixel\_y' represent the assigned area, and we assume the width and height of the images are both 100 for simplicity. Direction is chosen from `Left', `Backwards', `Forward' and `Right'.
    
    You can output an empty list if you believe there is no need for providing a new perspective.
    
    First think and then give the final answer. The final answer should contain a short summary of your reasoning process (no more than 200 words), which only focus on how to derive the command without giving the answer of the original question, and the final list.
    
    The summary should be put within $<$summary$>$ and $<$/summary$>$, and then output the list within $<$answer$>$ and $<$/answer$>$.

    \item \textbf{Reasoning prompt}:  First think and then give the final answer. Put the option letter within $<$answer$>$ and $<$/answer$>$.
    \item \textbf{Captioning prompt}: Describe the 3D location of the objects in the image. The location include left, right, top, bottom, background, foreground, and so on. You can describe the location of each object separately and the relative location between objects as well (e.g. Object A is to the right of Object B). The answer should be no more than 200 words.
\end{itemize}

\subsubsection{Data Sample}
We present a sample of \MethodName-128k in Fig.~\ref{fig:app-sftcase}. The sample contains the original problem input $(\mathcal{I}, q)$ and the ASM trajectory $(r_{\mathrm{plan}},\{C_{\mathrm{edit},k}\}_{k=1}^{K},r_{\mathrm{reason}},\hat{a})$. In this case, the number of auxiliary imagined view is $K=1$. Critically, the generated novel views only act as scaffolding for conversion into $\{C_{\mathrm{edit},k}\}_{k=1}^{K}$, and do not really exist in the final dataset.

\subsection{\MethodName-RL-58k}

\begin{figure}[h]
\centering
\includegraphics[width=1.0\textwidth]{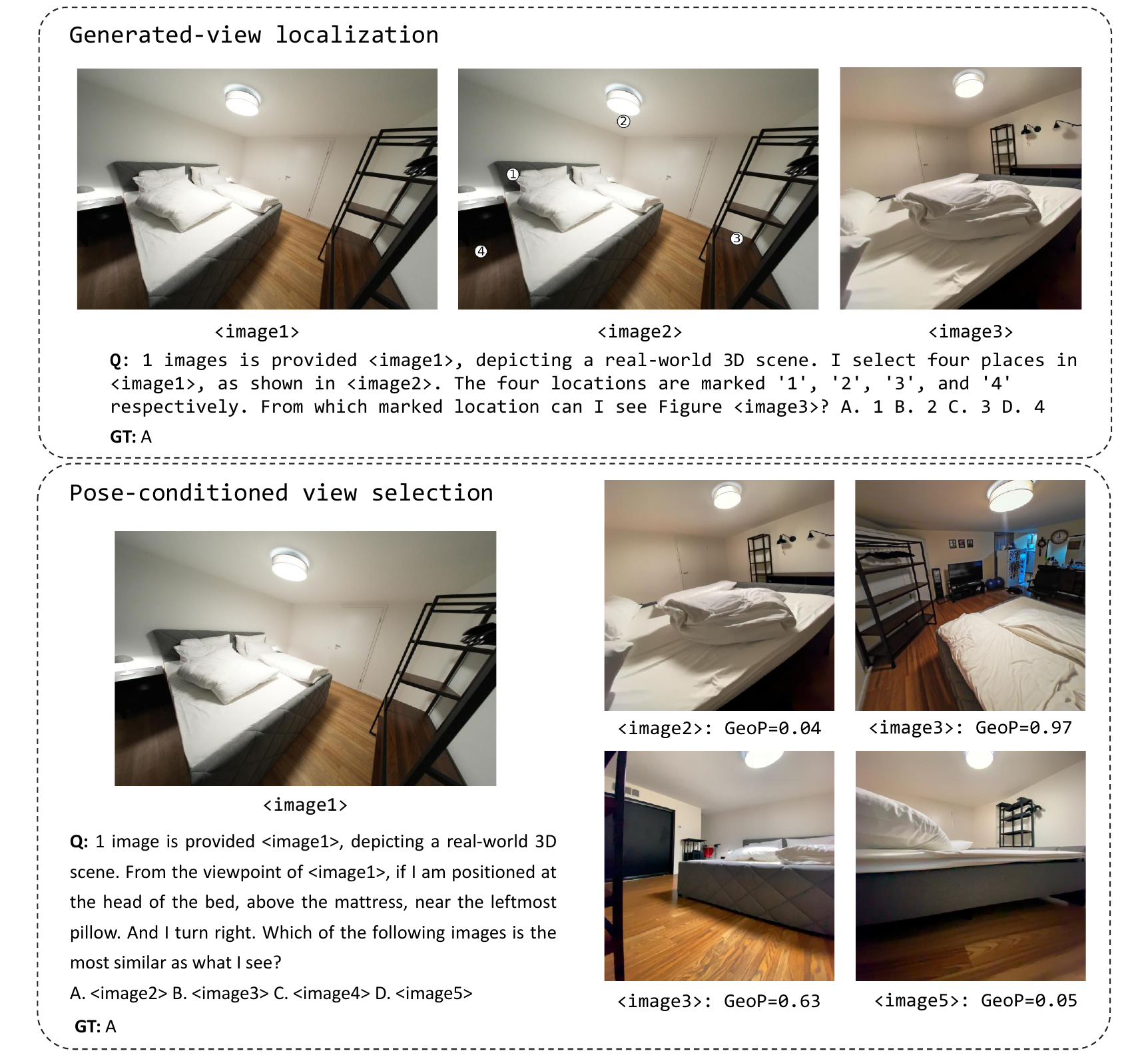}
\vspace{-12pt}
\caption{Examples of view-imagination RL tasks.}
\label{fig:app-rlcase}
\end{figure}

We construct the \MethodName-RL-58k automatically by employing \MethodName-NVG without any LVLM/human annotation. Only the original 2D images come from Holi-Spatial~\citep{gao2026holi} and DL3DV~\citep{ling2024dl3dv}. \MethodName-RL-58k consists of 58,346 training samples, with 28,594 from Holi-Spatial and 29,752 from DL3DV.

In Sec.~\ref{sec:asm-rl} we introduce the \textit{answer selection} strategy to reduce potential ambiguity in the two multi-choice tasks and ensure the quality of the rendered images. A pre-defined parameter $\theta_{RL}$ is applied in this strategy. During the construction process, we find that generated images generally have higher GeoP values when the queried allocentric viewpoint is set to \textit{backwards} than when it is set to other directions. This is because the backward direction is more likely to reveal regions not visible in the input images, requiring the NVG to synthesize more unobserved content. Therefore, even when the generated images exhibit similar levels of geometric consistency, those generated for the backward direction tend to receive higher GeoP. After manually inspecting a large number of generated images and their corresponding GeoP values, we set to $\theta_{RL}$ to $1.2$ for images whose corresponding target direction is \textit{backwards}, and 0.6 for all other directions.

In Fig.~\ref{fig:app-rlcase}, we give example VQAs of the two tasks we construct for RL: \textit{generated-view localization} and \textit{pose-conditioned view selection}. The formulation of both tasks are illustrated in Sec.~\ref{sec:asm-rl}.

\section{Experiment Details}
\subsection{Training Details}
\label{Appendix:training}
\textbf{\MethodName-NVG training.} We finetune FLUX.2-klein-9B~\citep{blackforestlabs2025flux2}, an image-editing model for real-time text-conditioned image generation, on \MethodName-edit140k to derive our \MethodName-NVG. Drawing on the training strategy in DiffThinker~\citep{he2025diffthinker}, we use mean squared error (MSE) as our training objective and train for 3 epochs. We only finetune its Diffusion Transformer (DiT) with Low-Rank Adaptation (LoRA) and we set LoRA rank to 512. For GeoP calculation, we use Mapanything~\citep{keetha2026mapanything} as the feed-forward 3D reconstruction model $F$ to reconstruct the 3D point clouds.

\textbf{View-imagination RL.} During GRPO, we apply KL regularization with weight 1e-2. And we rollout 5 responses in each group with the temperature set to 1.0. Our reward is composed of two items: accuracy reward and format reward, with the weight of 0.9 and 0.1. Accuracy reward is a binary reward which is assigned to 1 only when the final option is correct. Format reward is also binary which requires the LVLMs to reason first and put the option letter within $<$answer$>$ and $<$/answer$>$. 

\textbf{Prompt.} For both SFT and RL, we append a system prompt to the end of the original question, guiding LVLMs to output first reasoning content and then the final answer. The system prompt is: \textit{First think and then give the final answer. Put the option letter within $<$answer$>$ and $<$/answer$>$}.

\subsection{Benchmark Overview}
\label{Appendix:benchmark}

\begin{table}[t]
  \caption{Performance comparison on three general-VQA benchmarks.}
  \vspace{-12pt}
  \centering
  \setlength{\tabcolsep}{2.4pt}
  \normalsize
  \begin{tabular}{@{}lccc|c@{}}
    \toprule
    Models & $\text{MMMU}_\text{VAL}$ & MMStar & RealWorldQA & Average
      \\
    \midrule
    \rowcolor{gray!10}
    \multicolumn{5}{c}{\textit{Representative Spatial Models}} \\
    SpaceR-7B  & 51.8 & 61.4 & 65.0 & 59.4 \\
    VST-RL-7B  & 51.2 & 63.7 & 71.4 & 62.1  \\
    HoliSpatial-8B & 56.0 & 65.4 & 68.0 & 63.1 \\
    SenseNova-8B &  55.1 & 64.5 & 71.6 & 63.7  \\
    \midrule
    \rowcolor{gray!10}
    \multicolumn{5}{c}{\textit{Baselines \&  Our Model (Qwen3-VL-4B)}}\\
    Qwen3-VL-4B & 54.7 & 65.1 & 71.3 & 63.7 \\
    \MethodName-4B & 53.8 & 64.0 & 71.6 & 63.1 \\
    \midrule 
    \rowcolor{gray!10}
    \multicolumn{5}{c}{\textit{Baselines \&  Our Model (Qwen3-VL-8B)}}\\
    Qwen3-VL-8B & 54.1 & 66.0 & 73.2 & 64.4\\
    \MethodName-8B & 56.1 & 67.1 & 71.8 & 65.0 \\

\midrule 
    \rowcolor{gray!10}
    \multicolumn{5}{c}{\textit{Baselines \&  Our Model (Qwen3-VL-30BA3B)}}\\

    \small{Qwen3-VL-30BA3B} & 62.7 & 70.7 & 74.8 & 69.4  \\
    \small{\MethodName-30BA3B} & 62.1 & 68.6 & 76.1 & 68.9\\

    \midrule 
    \rowcolor{gray!10}
    \multicolumn{5}{c}{\textit{Baselines \&  Our Model (Qwen3.5-9B)}}\\
    
    Qwen3.5-9B & 73.3 & 78.6 & 82.4 & 78.1  \\
    \MethodName-9B & 75.3 & 78.5 & 81.7  & 78.5\\
    \bottomrule
  \end{tabular}
  \vspace{-8pt}
  \label{tab:app-gnrvqa}
\end{table}
We evaluate the LVLMs on \textbf{seven} spatial reasoning (SR) benchmarks. For Spatial457~\citep{wang2025spatial457}, we only select its most challenging 6d subtask with 1837 questions.  \textbf{ESI-Bench}~\citep{hong2026esi}, \textbf{MindCube}~\citep{wang2025spatial}, \textbf{MMSI-Bench}~\citep{yang2026mmsi}, and \textbf{Spatial457-6d}  are the most difficult among the seven benchmarks, requiring LVLMs to go beyond directly reading spatial relations from the given observations and instead integrate incomplete evidence, mentally construct a 3D scene representation, and reason about unobserved or hypothetical viewpoints. \textbf{ViewSpatialBench}~\citep{li2026viewspatial} represents an intermediate level of difficulty, with camera-perspective questions which can often be answered directly from the observed coordinate frame and person-perspective questions which require some basic perspective-taking capabilities. Finally, \textbf{3DSRBench}~\citep{ma20253dsrbench} and \textbf{EmbSpatial-Bench}~\citep{du2024embspatial} are relatively perception-oriented, covering spatial properties such as height, orientation, location, and  egocentric spatial relations. Taken together, these benchmarks provide a comprehensive evaluation of spatial reasoning across complementary dimensions, ranging from fundamental spatial perception and egocentric relation recognition to perspective transformation, multi-view scene reconstruction, and imagination-driven reasoning about unobserved spaces.

We additionally evaluate on \textbf{three} general-VQA benchmarks to examine whether \MethodName{} preserves the broad visual capabilities of the base LVLMs. \textbf{$\text{MMMU}_\text{val}$}~\citep{yue2024mmmu} evaluates expert-level multimodal understanding and reasoning across 6 major disciplines. \textbf{MMStar} \citep{chen2024we} evaluates 18 detailed ability axes in 6 major categories, spanning fine-grained and coarse visual perception, instance-level reasoning, and science and technology, with an emphasis on genuine visual dependency. \textbf{RealWorldQA} \citep{xai2024realworldqa} focuses on basic visual understanding in real-world scenes, including counting, object size comparison, relative direction, optical character recognition (OCR), etc. Together, these benchmarks measure whether the spatial improvements introduced by \MethodName{} come at the expense of general visual understanding.

\subsection{Evaluation Details}
\label{Appendix:evaluation}
\textbf{Evaluation prompt.} In Tab.~\ref{tab:main}, we evaluate the LVLMs on two major task categories: spatial reasoning and general-VQA (Gnr.-VQA). For spatial reasoning, we apply the same system prompt as training (in Appendix~\ref{Appendix:training}) to let LVLMs reason before giving the answer. For general-VQA benchmarks, we apply another system prompt to enable direct responses of final answers from LVLMs: \textit{Directly output your answer without any reasoning. The correct option letter is:}.

\textbf{Representative spatial models.} Four popular spatial LVLMs of previous studies are also evaluated on the benchmarks. SpaceR-7B~\citep{ouyang2025spacer} and VST-RL-7B~\citep{yang2026visual} are trained upon Qwen2.5-VL-7B~\citep{qwen2.5-VL}. Their official repository provides system prompts. For the SR evaluation, we follow them exactly to guide the models to produce structured reasoning and final answers. Holi-Spatial~\citep{gao2026holi} and 
SenseNova-SI-1.1-Qwen3-VL-8B~\citep{cai2026scaling} are trained upon Qwen3-VL-8B~\cite{bai2025qwen3}, similar to out \MethodName-8B. Since the two models do not provide official system prompt, we follow our system prompt during evaluation.

\subsection{General VQA Results}
\label{Appendix:general}
In Tab.~\ref{tab:main}, we only report the average score on the three general-VQA benchmarks. We provide the detailed accuracy of all LVLMs on each benchmark in Tab.~\ref{tab:app-gnrvqa}.

\subsection{Ablation Details}
\label{Appendix:ablation}

In Fig.~\ref{fig:explicit1} and Tab.~\ref{tab:comp-NVG}, we employ the reasoning paradigm of external ASM: LVLMs execute question-conditioned view-planning, then invoke the NVG to synthesize corresponding observations, and finally reason based on the previous auxiliary information. Three different LVLMs are evaluated in Fig.~\ref{fig:explicit1}: Qwen3-VL-8B~\citep{bai2025qwen3}, Qwen3.5-35BA3B~\citep{qwen36_35b_a3b}, and Kimi-k2.5~\citep{team2026kimi}. And four different rendering models are compared in Tab.~\ref{tab:comp-NVG}: GPT-Image-2~\citep{openai2026gptimage2}, SpatialEdit~\citep{xiao2026spatialedit}, SVC~\citep{zhou2025stable}, and our \MethodName-NVG.

In Tab.~\ref{tab:comp-NVG}, GeoP is used for comparing the synthesis quality of different NVGs. For each setting, we report the median GeoP of the images generated by the NVG rather than the average. This is because most generated images have low GeoP values (below 1), while a small number of poor-quality images have much higher values (above 10). These outliers can disproportionately raise the average GeoP, making it less representative of the geometric consistency of a typical novel-view observation generated by a certain NVG/World-Model.

Tab.~\ref{tab:abl} ablates training and data-filtering strategies described in Sec.~\ref{sec:asm-sft} and Sec.~\ref{sec:asm-rl}. For SFT, \textit{w/o filter} means omitting the trajectory \textbf{filtering} step in Sec.~\ref{sec:asm-sft} and using all synthesized ASM trajectories for training. For RL, \textit{w/o filter} means omitting the GeoP-based criterion in the \textbf{answer selection} step of Sec.~\ref{sec:asm-rl}: no sample is discarded because its candidate views exceed the GeoP threshold, and the ground-truth candidates are selected from the full candidate set. Importantly, the final \textbf{SFT+RL} setting in Tab.~\ref{tab:abl} as well as all our models shown in Tab.~\ref{tab:main} experience both training stages and all training data experience the filtering strategy.

\section{More Cases}
\subsection{\MethodName-NVG Rendering}

\begin{figure}[h]
\centering
\includegraphics[width=1.0\textwidth]{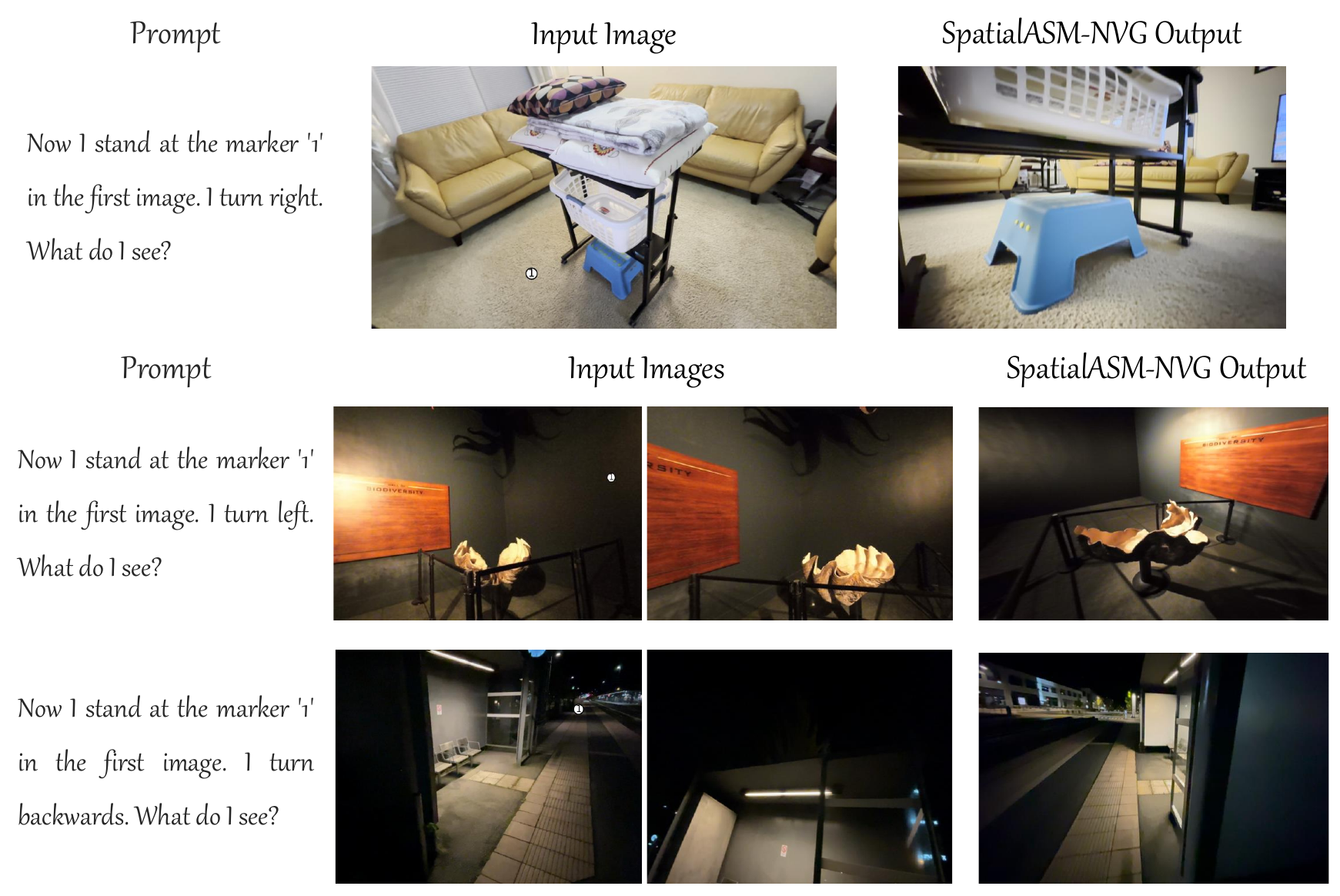}
\vspace{-12pt}
\caption{Examples of novel-view observations rendered by \MethodName-NVG.}
\label{fig:app-nvgcase}
\end{figure}

We obtain \MethodName-NVG after training FLUX.2-klein-9B~\citep{blackforestlabs2025flux2} on \MethodName-edit140k. Here we provide three output examples rendered by \MethodName-NVG in Fig.~\ref{fig:app-nvgcase}.

\textit{Case 1:} Turning right at the target location reveals an area that was originally outside the input image’s field of view (beyond the sofa on the far right). \MethodName-NVG successfully reconstructs \textbf{the missing content} in this region while retaining the objects in the original input (e.g., blue chair, white box). 

\textit{Case 2:} The generated view is not only geometrically consistent with the input but also faithfully reproduces the \textbf{lighting} on the brown panel on the right and the \textbf{shadows} cast on the floor.

\textit{Case 3:} This case is indeed rather tricky, as the relative camera poses of the two input images are not immediately apparent. Nevertheless, our model correctly infers their spatial relationship and \textbf{seamlessly integrates} the tactile paving from Image 1 (invisible in Image 2) with the white panel on the inner left side of the bus stop from Image 2 (not visible in Image 1) into a single rendered image.

\subsection{Results of Novel-view Generators}
\begin{figure}[h]
\centering
\includegraphics[width=1.0\textwidth]{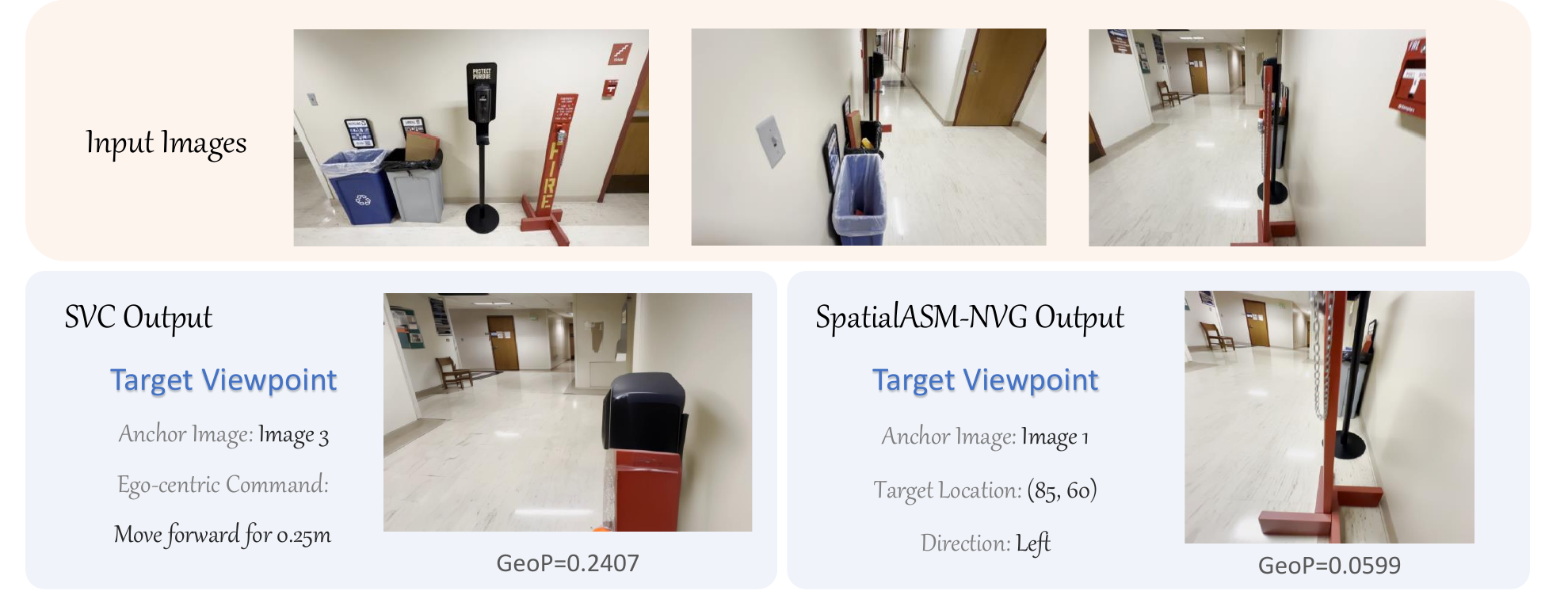}
\vspace{-12pt}
\caption{Case 1: Observations generated by different novel-view generators.}
\label{fig:app-compcase1}
\end{figure}

\begin{figure}[h]
\centering
\includegraphics[width=1.0\textwidth]{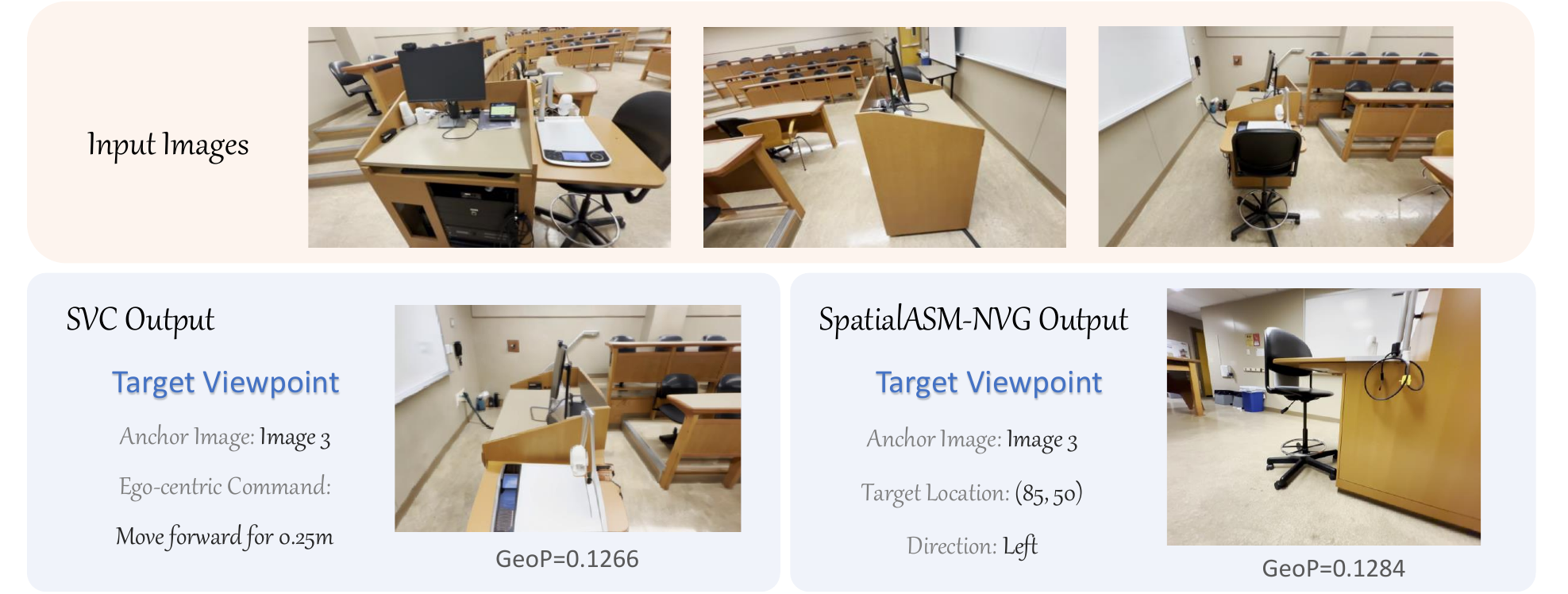}
\vspace{-12pt}
\caption{Case 2: Observations generated by different novel-view generators.}
\label{fig:app-compcase2}
\end{figure}

To showcase qualitative examples of the rendering results of different novel-view generators, we select some generated observations from the experiment in Tab.~\ref{tab:comp-NVG} with Kimi-k2.5 as the LVLM for planning and reasoning. The generated images are shown in Fig.~\ref{fig:app-compcase1} and Fig.~\ref{fig:app-compcase2}. Evidently, the formulation of target viewpoints varies greatly. Prior approaches(e.g., SVC~\citep{zhou2025stable}) use ego-centric motions from the camera position of an anchor image, whereas \MethodName-NVG defines the target location directly by marking a pixel on an image. Note that in both figures we assume that the input images all have the size $(100, 100)$.

\textit{Case 1 (Fig.~\ref{fig:app-compcase1}):} The query target viewpoints of the two novel-view generators are approximately the same. \MethodName-NVG has a much lower GeoP value because the red fire-fighting unit and the black disinfectant unit in the output image more closely resemble their counterparts in the input images. 

\textit{Case 2 (Fig.~\ref{fig:app-compcase2}):} Although SVC’s output image is more geometrically consistent with the visual inputs (lower GeoP), its egocentric formulation results in relatively small camera motions and its output remains highly similar to the anchor image, providing little additional information for further LVLM reasoning. In contrast, our allocentric formulation induces larger viewpoint changes and provides more informative output observations, which could also possibly help the LVLM better understand the relative camera poses across input images and the spatial layout of the 3D scene.

\subsection{Examples of LVLM Internalization }

\begin{figure}[h]
\centering
\includegraphics[width=1.0\textwidth]{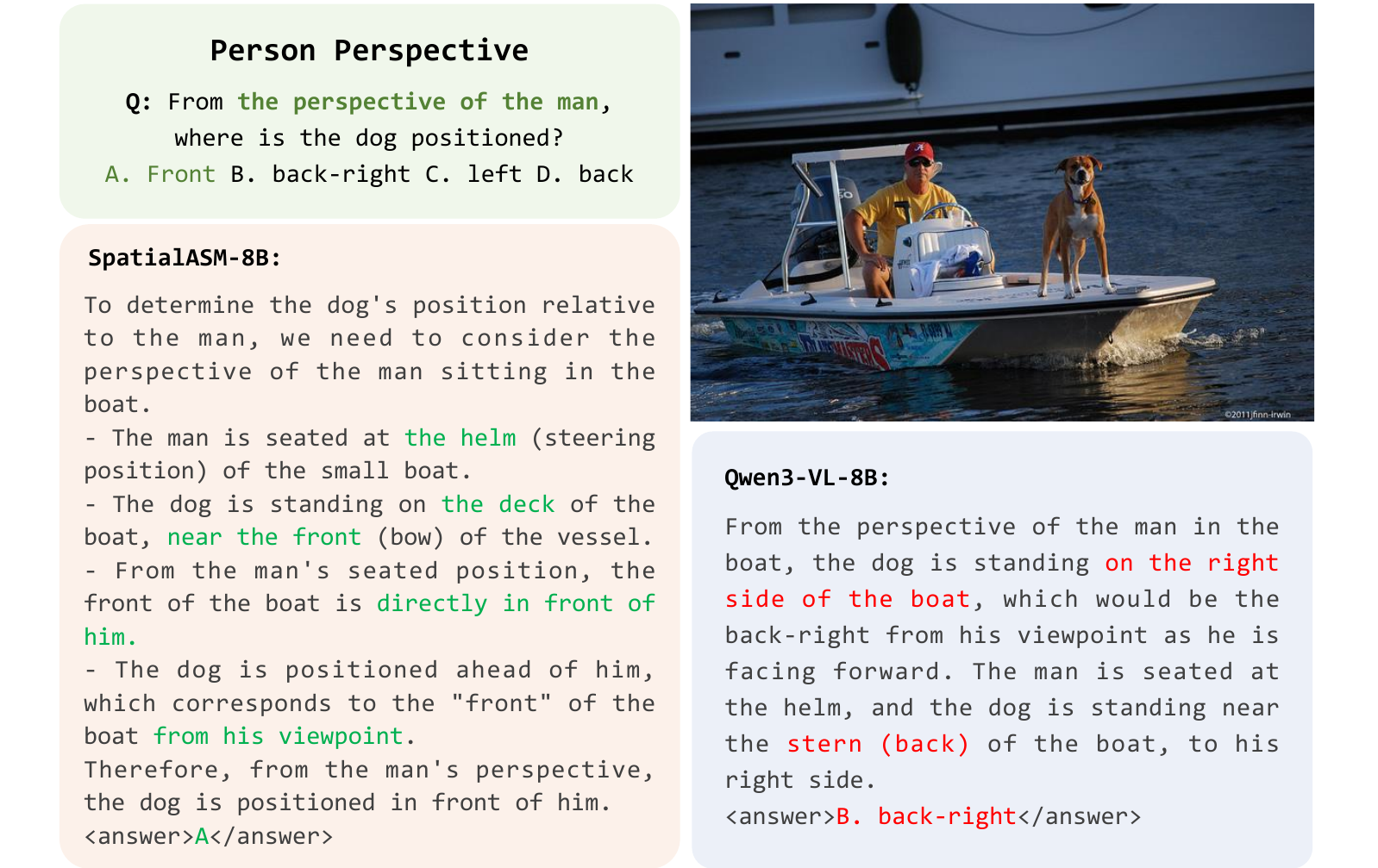}
\vspace{-12pt}
\caption{Qualitative comparison of Qwen3-VL-8B and \MethodName-8B.}
\label{fig:app-VLMcase1}
\end{figure}

We have trained several LVLMs to internalize ASM capabilities and improved their overall spatial reasoning performance (Tab.~\ref{tab:main}). Here, we provide \textbf{four qualitative comparisons} between \MethodName-8B and its baseline, Qwen3-VL-8B. These examples show that \MethodName-8B can adaptively decide whether to employ the ASM reasoning pattern in its output according to the task type and difficulty, rather than simply overfitting to the ASM reasoning trajectories. In fact, we find that the model does not explicitly follow the ASM pattern in most cases, yet its overall spatial reasoning accuracy still improves substantially (Tab~\ref{tab:main}). This suggests that our training \textbf{internalizes generalizable spatial reasoning capabilities} instead of imposing a rigid reasoning template.

\begin{figure}[h]
\centering
\includegraphics[width=1.0\textwidth]{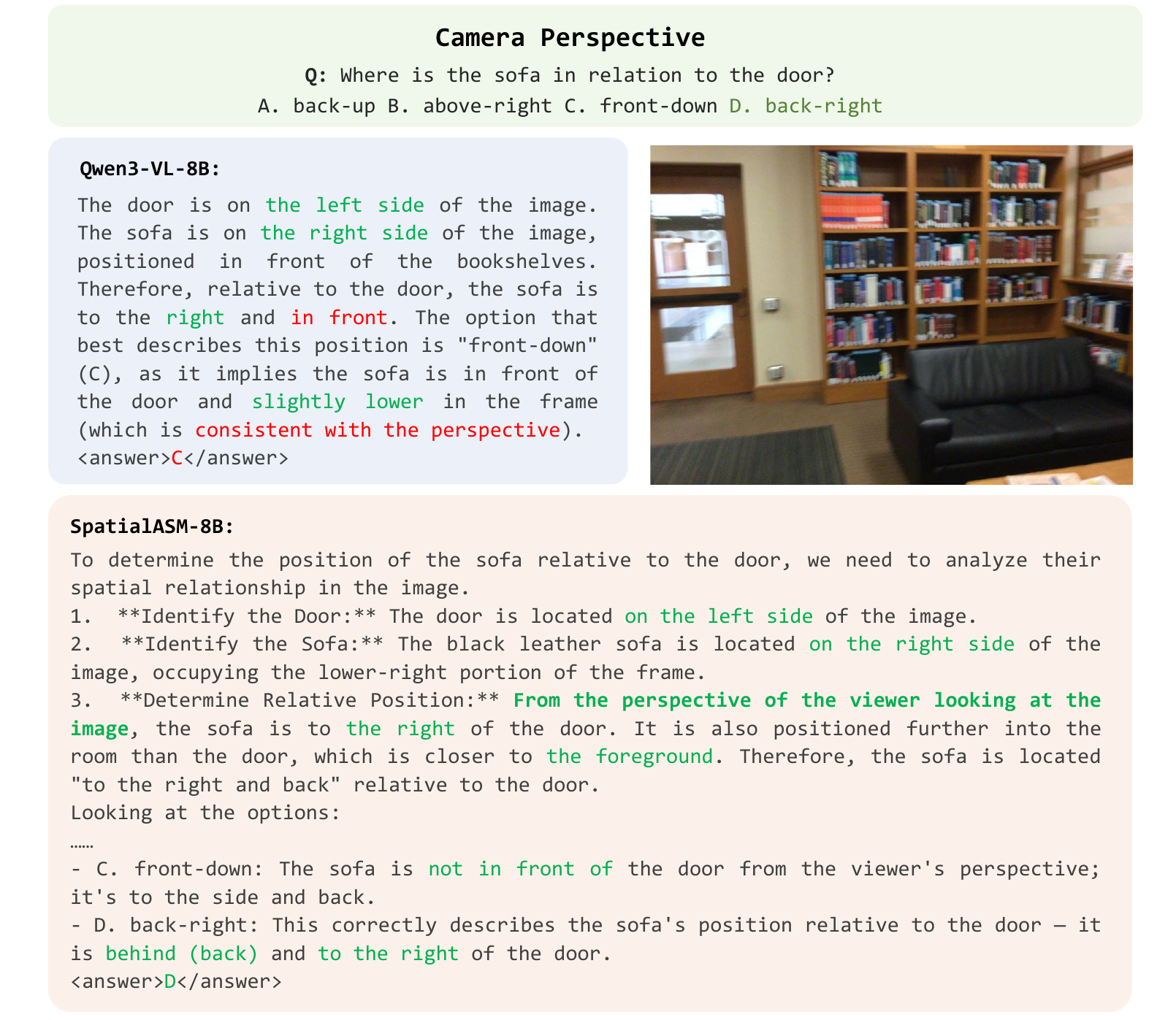}
\vspace{-12pt}
\caption{Qualitative comparison of Qwen3-VL-8B and \MethodName-8B.}
\label{fig:app-VLMcase2}
\end{figure}

\begin{figure}[h]
\centering
\includegraphics[width=1.0\textwidth]{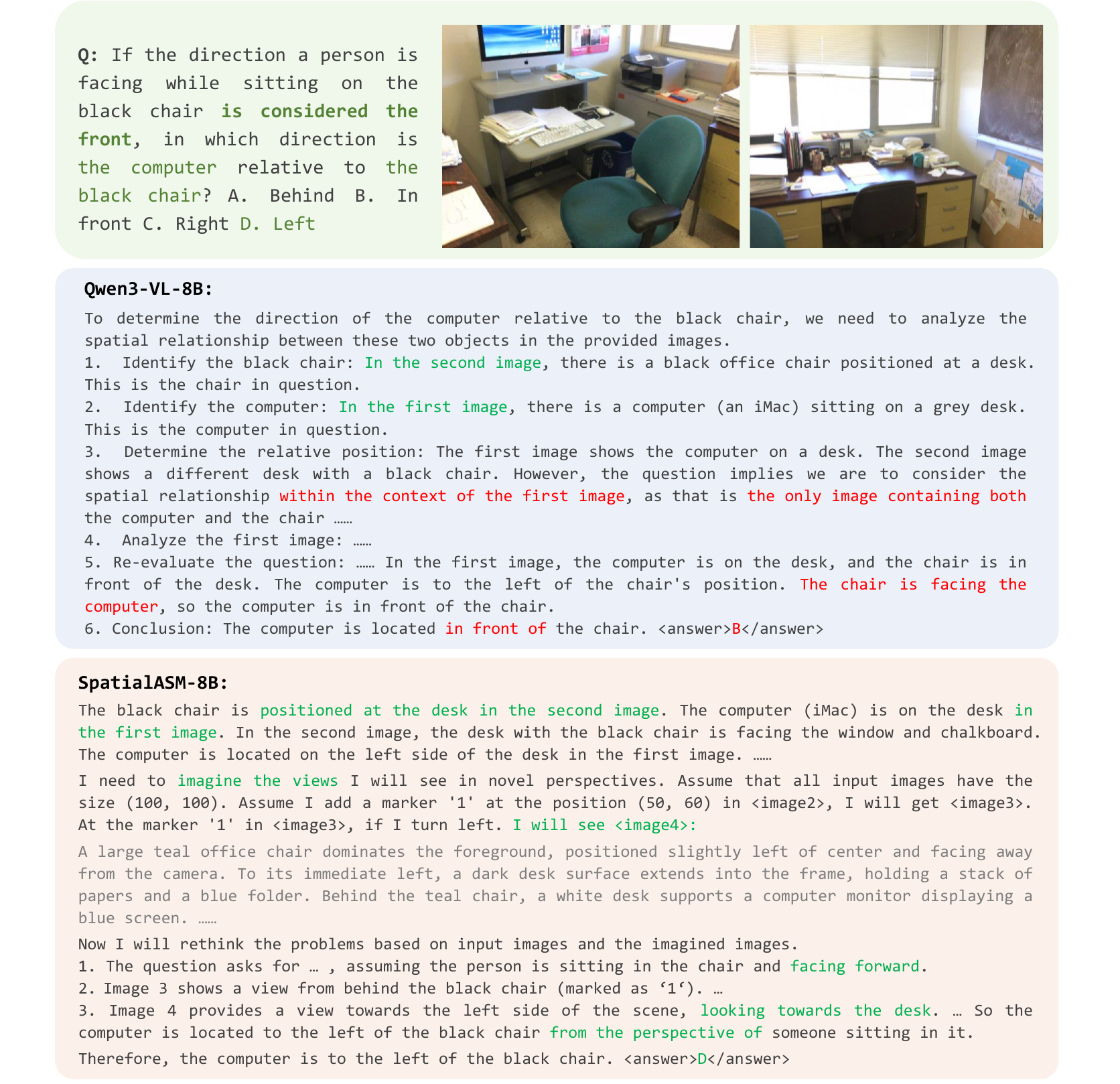}
\vspace{-12pt}
\caption{Qualitative comparison of Qwen3-VL-8B and \MethodName-8B.}
\label{fig:app-VLMcase3}
\end{figure}

\begin{figure}[h]
\centering
\includegraphics[width=1.0\textwidth]{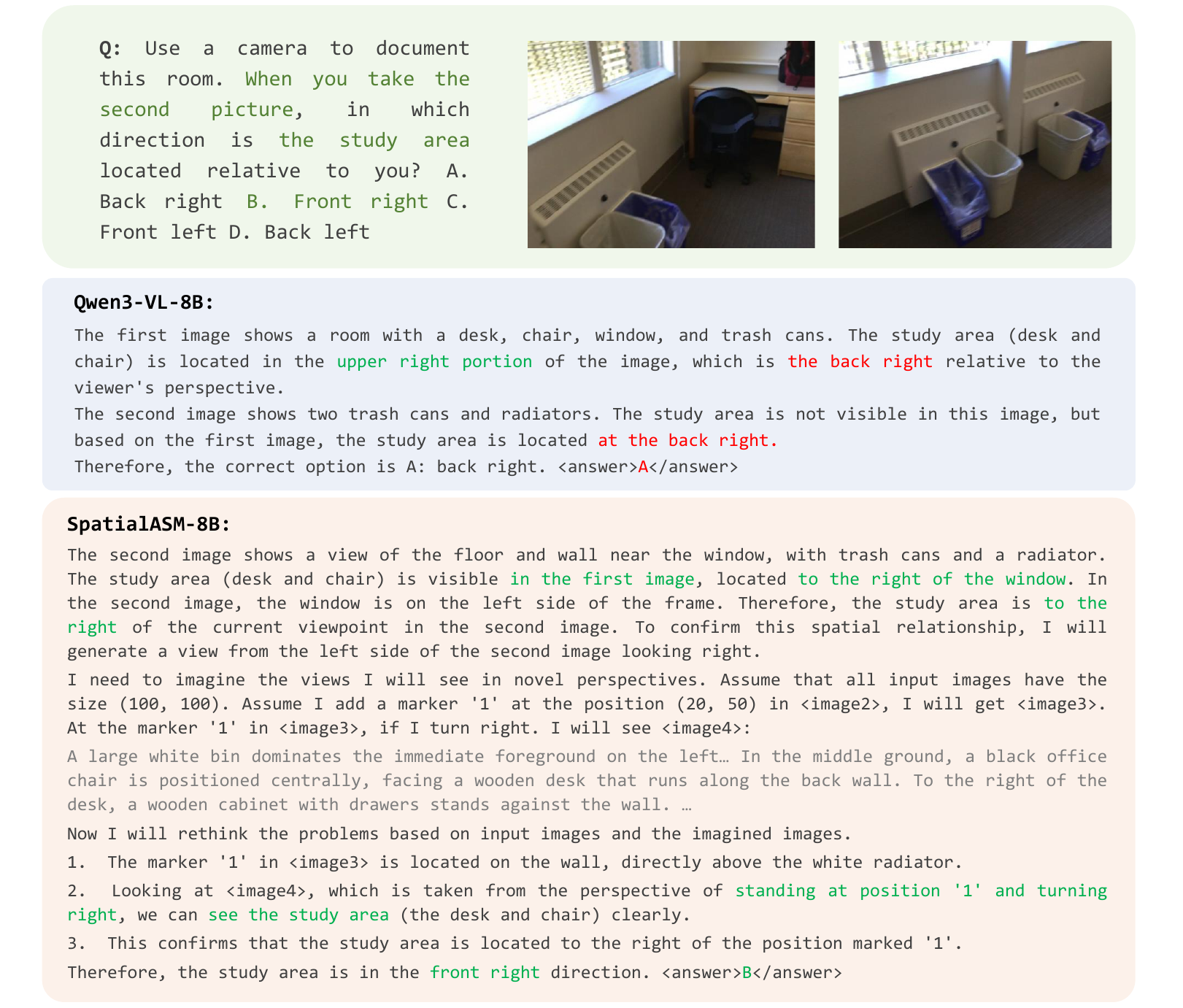}
\vspace{-12pt}
\caption{Qualitative comparison of Qwen3-VL-8B and \MethodName-8B.}
\label{fig:app-VLMcase4}
\end{figure}

Fig.~\ref{fig:app-VLMcase1} and Fig.~\ref{fig:app-VLMcase2} show two problems targeting the viewpoint of \textit{person} and \textit{camera} respectively. Though \MethodName-8B does not produce the ASM reasoning pattern explicitly, it accurately interprets the 3D spatial configuration from different perspectives and correctly answers both questions. Fig.~\ref{fig:app-VLMcase3} and Fig.~\ref{fig:app-VLMcase4} are more challenging tasks where \MethodName-8B generates CoT of  active spatial planning and imagination to help reach the correct answer.

\end{document}